\documentclass{article}

\PassOptionsToPackage{numbers, compress}{natbib}
\usepackage[main, final]{neurips_2025}

\usepackage[utf8]{inputenc} % allow utf-8 input
\usepackage[T1]{fontenc}    % use 8-bit T1 fonts
\usepackage{hyperref}       % hyperlinks
\usepackage{url}            % simple URL typesetting
\usepackage{booktabs}       % professional-quality tables
\usepackage{amsfonts}       % blackboard math symbols
\usepackage{nicefrac}       % compact symbols for 1/2, etc.
\usepackage{microtype}      % microtypography
\usepackage{xcolor}         % colors
\usepackage{amssymb}
\usepackage{amsmath}
\usepackage{amsthm}
\usepackage{algorithm}
\usepackage{algpseudocode}
\usepackage{tikz}
\usepackage{caption}
\usepackage{tcolorbox}
\usepackage{pgfplots}
\usepackage{array}
\usepackage{adjustbox}
\usepackage{wrapfig}
\usepackage{xspace}
\usepackage{listings}
\usepackage{makecell}
\usepackage{tabularx, array}
\usepackage{enumitem}
\usepackage{mathrsfs}  % in preamble

\usepackage{capt-of}      % for \captionof
\usepackage{subcaption}

\newcolumntype{L}{>{\raggedright\arraybackslash}p{0.16\textwidth}} % just wide enough for “Surprisingness”
\newcolumntype{F}{>{\raggedright\arraybackslash}p{0.19\textwidth}} % wraps long math nicely

\PassOptionsToPackage{frozencache,cachedir=.}{minted}
\usepackage{minted}
\usepackage[normalem]{ulem}
\usepackage{appendix}
\usepackage{etoc}
\usepackage{comment}
\usepackage[symbol]{footmisc}
\usepackage{pgfplots}
\pgfplotsset{width=10cm,compat=1.15}
\usepackage{mdframed}

\usepackage[textsize=scriptsize]{todonotes}
\usepackage{cleveref}

\newcommand{\method}{EvoAbstract\xspace}
\newcommand{\fermat}{\textsc{Fermat}\xspace}
\usepackage{thmtools} % Optional, enhances amsthm

\newcommand{\forceindent}{\leavevmode{\parindent=1em\indent}}
\newcommand{\bx}{\mathbf{x}}
\newcommand{\by}{\mathbf{y}}
\newcommand{\bp}{\mathbf{p}}
\newcommand{\bv}{\mathbf{v}}
\newcommand{\bm}{\mathbf{m}}
\newcommand{\lam}{\lambda}

\newcommand{\cA}{\mathcal{A}}
\newcommand{\cB}{\mathcal{B}}
\newcommand{\cF}{\mathcal{F}}
\newcommand{\cG}{\mathcal{G}}
\newcommand{\cH}{\mathcal{H}}
\newcommand{\cP}{\mathcal{P}}
\newcommand{\cQ}{\mathcal{Q}}
\newcommand{\cR}{\mathcal{R}}

\theoremstyle{definition}
\theoremstyle{plain} % For conjectures and theorems

\theoremstyle{remark} % For proofs, remarks, and informal text

\title{Learning Interestingness in Automated Mathematical Theory Formation}

\author{
  George Tsoukalas \\
  UT Austin \\
  george.tsoukalas@utexas.edu
  \And
  Rahul Saha \\
  UT Austin \\
  rahul.saha@utexas.edu
  \And
  Amitayush Thakur \\
  UT Austin \\
  amitayush@utexas.edu
  \And
  Sabrina Reguyal \\
  Princeton University, Stanford University \\
  sreguyal@stanford.edu
  \And
  Swarat Chaudhuri \\
  UT Austin \\
  swarat@cs.utexas.edu
}

\begin{document}

\maketitle

\begin{abstract}
We take two key steps in automating the open-ended discovery of new mathematical theories, a grand challenge in artificial intelligence. First,
we introduce \fermat, a reinforcement learning (RL) environment that models concept discovery and theorem-proving using a set of symbolic actions, opening up a range of RL problems relevant to theory discovery.
Second, we explore a specific problem through \fermat: automatically scoring the \emph{interestingness} of mathematical objects. We investigate 
evolutionary algorithms for synthesizing nontrivial interestingness measures. In particular, we introduce an LLM-based evolutionary algorithm that features function abstraction, leading to notable improvements in discovering elementary number theory and finite fields over hard-coded baselines. We open-source the \fermat environment at \href{https://github.com/trishullab/Fermat}{{\texttt{github.com/trishullab/Fermat}}}.
\end{abstract}

\section{Introduction}
AI researchers have dreamed of building an ``automated mathematician'' since the 1950s \citep{newell1956logic}. 
Such a system would allow human mathematicians to harness the vast processing capacity of computers to discover entirely new areas of mathematics \citep{shulman2024-ams-strangenew}.
An emerging body of work seeks to realize this dream using the tools of modern machine learning. In particular, the AI community has developed a wide range of systems that can prove formal theorems \citep{deepmind2024alphaproof, trinh2024solving} and search for programs discovering mathematical constructions \citep{AlphaEvolve, romera-paredes2023mathematical}.

However, a key limitation of much of this research is that it is focused on solving predefined problems. 
Mathematicians develop theories through an \emph{open-ended} process of defining new concepts, studying their properties, making conjectures, and proving or finding counterexamples. While some work \citep{poesia2024learningformalmathematicsintrinsic} has offered systems that construct new problems in addition to solving them, there is currently no framework that supports the full theory-formation process, including, for example, the synthesis of new definitions in addition to problems. 

A central challenge in this open-ended process is guiding the search. The space of possible definitions and conjectures is combinatorially vast, and most paths lead to trivial or dull mathematics. Human mathematicians navigate this space using a nuanced, intuitive sense of ``interestingness" --- a judgment of scientific potential that directs their focus. An explicit formulation of this concept has long been debated, with different perspectives valuing properties such as the surprising connection between disparate fields \citep{Poincare1910-POIMC}, depth and generality \citep{Hardy1941-HARAMA-15}, or its unexpected real-world applicability \citep{wigner1960}.

\begin{wrapfigure}{r}{0.5\textwidth}
  \centering
  \vspace{-16pt}
  \includegraphics[width=0.5\textwidth]{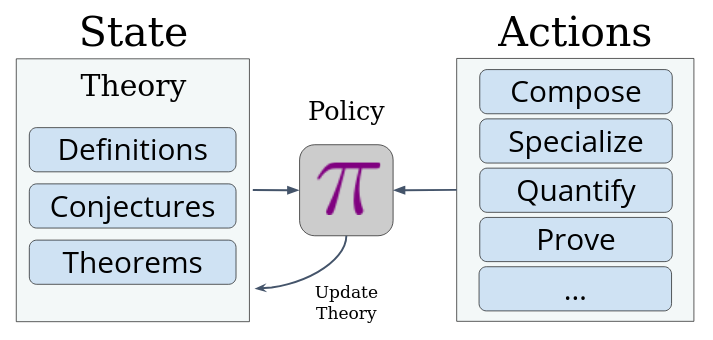}
 % \vspace{-10pt}
  \caption{A high-level description of \fermat, our environment for mathematical theory formation. At any given time, the current theory (state) is represented as a knowledge graph consisting of the mathematical definitions, conjectures, and theorems discovered so far. At each step, the policy $\pi$ inputs the current state and selects an action to apply, updating the theory with additional information. The action space allows the production of new definitions, conjectures, and proofs of theorems.}
  \vspace{-13pt}
  \label{fig:fermat}
\end{wrapfigure}

In this paper, we take two key steps towards addressing these challenges. First, we provide a reinforcement learning (RL) framework, called \fermat (\Cref{fig:fermat}), which can be used to design and evaluate new algorithms for automatic theory formation. 
The system generalizes the early symbolic computing-prover system HR \citep{colton2000hr}, which used a system of \emph{production rules} to generate new concepts and conjectures, either symbolically or from explicit examples, and \emph{proof mechanisms} for resolving conjectures. We model these symbolic steps as the actions of a Markov Decision Process (MDP), and the mathematical knowledge available at a given point during exploration as an MDP state. This formulation opens up numerous RL problems relevant to theory formation.

Our second contribution is a solution to a particular algorithmic problem in \fermat: learning an interestingness heuristic for selecting mathematical concepts to develop. To form a theory, one must navigate a combinatorial search space of mathematical objects, most objects in which are not meaningful or worthy of study. Prior works were attentive to this problem, but required hard-coded measures to formalize the concept of interestingness \citep{colton2000hr, lenat1977am}. In contrast, we frame the discovery of this heuristic as a learning problem. We specifically learn interestingness measures as \emph{programmatic} representations, as this makes them interpretable and allows us to analyze what may contribute to fruitful discovery. To this end, we develop an LLM-driven method, called \method, for learning the intrinsic value of mathematical objects in the context of the current theory. \method is an evolutionary program synthesis algorithm that extends the FunSearch \citep{romera-paredes2023mathematical} approach with a form of abstraction learning, allowing for interpretable abstractions to be discovered during function search. We experimentally show that \method can automatically synthesize interestingness measures that lead to significant improvements in discovering concepts in elementary number theory and finite fields over hard-coded baselines. 

\section{Problem Formulation and Motivation}
\subsection{Mathematical Theory Formation as a Markov Decision Process (MDP)}

To rigorously study automated mathematical theory formation using reinforcement learning, we first formalize the process as an MDP $(\mathcal{S}, \mathcal{A}, \mathcal{T}, \mathcal{R})$. This framework allows us to model the sequential nature of mathematical discovery, where an agent iteratively expands a body of knowledge by making choices about definitions, conjectures, and proof attempts. Let $\mathcal{M}$ denote the universe of all well-formed mathematical entities. The components of this MDP are defined as follows:

% one‑off bullet paragraph
% \hangindent=1.6em      % ← entire paragraph indented
\hangafter=0           %   from the first line
\noindent              %   (no extra left skip)
\textbullet\hspace{0.4em}
 \textbf{Mathematical State Space ($\mathcal{S}$):} A state $S \in \mathcal{S}$ represents the current state of mathematical knowledge, represented as a \emph{directed knowledge graph} $G = (V, E)$, where:
    \begin{itemize}[leftmargin=2em, labelsep=0.4em, nosep]
        \item $V \subseteq \mathcal{M}$ is the set of \textbf{mathematical entities}, categorized into definitions $\mathcal{D}$, conjectures $\mathcal{C}$, and theorems $\mathscr{T}$.
        \item $E$ is the set of dependency \textbf{edges},  where an edge $(u,v)$ exists if entity $u$ was used as direct input for the action that generated entity $v$, and is labeled with that action.
    \end{itemize}

\smallskip

% another bullet paragraph
% \hangindent=1.6em
\hangafter=0
\noindent
\textbullet\hspace{0.4em}%
\textbf{Action Space ($\mathcal{A}$):} An action $a \in \mathcal{A}$ represents an operation that modifies the knowledge graph by introducing a new entity or acting upon existing ones. Actions fall into the following categories:
    \begin{itemize}[leftmargin=2em, labelsep=0.4em, nosep]
        \item \textbf{Definition Production Actions ($\mathcal{A}_{def}$)}: Introduces a new definition, adding a node $d'$ to $G$ and connecting it to relevant entities via a function $\delta_{def}: \mathcal{S} \times \mathcal{A}_{def} \to \mathcal{S}$.
        \item \textbf{Conjecture Production Actions ($\mathcal{A}_{conj}$)}: Formulates a new conjecture $c'$ based on existing entities and relationships, governed by a function $\delta_{conj}: \mathcal{S} \times \mathcal{A}_{conj} \to \mathcal{S}$.
        \item \textbf{Proof Actions ($\mathcal{A}_{proof} = \{\textit{prove, disprove}\}$)}: Verifies or refutes a conjecture $c \in \mathcal{C}$ by invoking a backend theorem prover, updating its status to theorem or disproven.
    \end{itemize}

\smallskip

% another bullet paragraph
% \hangindent=1.6em
\hangafter=0
\noindent
\textbullet\hspace{0.4em}%
\textbf{Transition Function ($\mathcal{T}$):} The transition function $\mathcal{T}: \mathcal{S} \times \mathcal{A} \times \mathcal{S} \to [0,1]$ models how applying an action updates the knowledge graph. $\mathcal{T}(S, a, S')$ denotes the probability of transitioning from state $S$ to $S'$ after applying action $a$. In particular,
    \begin{itemize}[leftmargin=2em, labelsep=0.4em, nosep]
        \item Adding a new definition or conjecture $c$ extends $V$ and introduces edges emanating from the entities to which the production rule was applied: $V' \!=\! V \cup \{c\}$, $E' \!= \! E \cup \{(v, c) \mid v \in V_{inputs}\}$. %\george{Should this actually be a hypergraph? THe implementation isn't like that but I guess this is a 3-graph.}
        \item A successful prove action converts a conjecture $c$ into a theorem and attaches a proof attribute: $\mathcal{C}' = \mathcal{C} \setminus \{c\}$, $\mathscr{T}' = \mathscr{T} \cup \{t\}$ with proof structure $\pi_t$. A successful disprove action refutes the conjecture $c$, marking it as false and attaching a counterexample as a witness, where possible.
    \end{itemize}

\smallskip

% another bullet paragraph
% \hangindent=1.6em
\hangafter=0
\noindent
\textbullet\hspace{0.4em}%
\textbf{Reward Function:} We design an extrinsic reward function $\mathcal{R}_\mathcal{E}: \mathcal{S} \times \mathcal{A} \times \mathcal{S} \to \mathbb{R}$ to incentivize the discovery of a pre-defined set $\mathcal{E}$ of well-known mathematical entities. Let the application of action $a$ to state $S$ produce a state $S'$ with a new entity $m_{new} \in \mathcal{M}$. The reward is defined as:
\[
\mathcal{R}_{\mathcal{E}}(S, a, S') = 
\begin{cases} 
    1 & \text{if } m_{new} \in \mathcal{E} \\ 
    0 & \text{otherwise} 
\end{cases}
\]
A reward is thus granted only when the agent's action results in the discovery of a specific ground-truth concept. A \emph{policy}, denoted by $\pi(a|s)$, defines a strategy by specifying the probability of taking action $a$ in a given state $s$. A \emph{rollout} refers to a single episode of interaction used to evaluate this policy by generating a \textit{trajectory}, $\tau = (S_0, a_0, r_1, S_1, a_1, r_2, \dots, a_{T-1}, r_T, S_T)$. This sequence is formed by starting in $S_0$ and repeatedly sampling an action $a_t \sim \pi(\cdot|S_t)$, after which the environment dictates the next state $S_{t+1} \sim \mathcal{T}(S_t, a_t, \cdot)$ and the corresponding reward $r_{t+1} = \mathcal{R}_{\mathcal{E}}(S_t, a_t, S_{t+1})$.

The \emph{intrinsic reward} $\mathcal{R}_{\mathcal{I}}$ is a function $\mathcal{R}_{\mathcal{I}}: \mathcal{S} \times \mathcal{A} \times \mathcal{S} \to \mathbb{R}$ that serves as a mechanism for the agent/policy to learn effectively in a sparse extrinsic reward setting. Such internal rewards can be critical for driving exploration and acquisition of general knowledge about the environment and discovery of useful subgoals, especially when external feedback is infrequent or absent, by promoting behaviors like curiosity or novelty-seeking \citep{alet2020metalearningcuriosityalgorithms, bellemare2016unifyingcountbasedexplorationintrinsic, Oudeyer2007, pathak2017curiositydrivenexplorationselfsupervisedprediction, SchmidhuberIntrinsic, intrinsicRLevolution}.

\subsection{Interestingness as Intrinsic Reward}
Humans use intuition and are intrinsically motivated to define interesting mathematical goals. Capturing this notion for an autonomous agent is a key challenge. We approach this by modeling interestingness as a learnable intrinsic reward, guiding a policy to discover meaningful theory. In this work, we wish to discover such interestingness measures autonomously, and view the synthesis of an effective interestingness measure as a problem of \emph{intrinsic reward optimization}.

Formally, we define the interestingness measure to be a function $\mathcal{I} : \mathcal{M} \times \mathcal{S} \to \mathbb{R}$, where $\mathcal{I}(m, S)$ provides an estimate of the value of a mathematical entity $m$ in the context of the current theory $S$. We connect this entity-scoring function to our RL framework by defining the intrinsic reward $\mathcal{R}_{\mathcal{I}}$ for a state transition as the interestingness score of the newly generated entity. If taking action $a$ in state $S$ produces a new entity $m_{\text{new}}$ in state $S'$, the intrinsic reward is
$ \mathcal{R}_{\mathcal{I}}(S, a, S') = \mathcal{I}(m_{\text{new}}, S')$.

In this work, we search over the class of interestingness measures that can be represented as programmatic functions. The policy, $\pi_{\mathcal{I}}$, is built using the measure $\mathcal{I}$ according to a fixed template (detailed in Section 5), and is designed to leverage the function's scores to guide its selection of actions. While the policy $\pi_{\mathcal{I}}$ acts based on this local, short-term measure, our learning problem is to discover an optimal function $\mathcal{I}^*$ that maximizes the cumulative long-term \emph{extrinsic} reward:
$$ \mathcal{I}^* = \arg\max_{\mathcal{I}} \mathbb{E}_{\tau \sim \pi_{\mathcal{I}}} \left[ \sum_{t} \gamma^t \mathcal{R}_{\mathcal{E}}(S_t, a_t) \right] $$
where $\gamma \in [0,1]$ represents the discount factor. In Section \ref{seal}, we detail our evolutionary algorithm designed towards discovering an optimal $\mathcal{I}$.

\section{\fermat: A Framework for Automated Theory Formation}
In this section, we discuss \fermat, our framework for automated mathematical theory formation built atop our MDP formulation of the theory discovery. 

\subsection{Mathematical Entities}

The \textsc{Fermat} framework, implemented in Python, provides the environment for automated theory formation. It is built upon a structured representation of mathematical entities within an evolving knowledge graph. At its core is a formal domain specific language (DSL) to define these entities. 

Each mathematical entity $m$ within the knowledge graph $G=(V,E)$ (where $m \in V$) encapsulates its meaning through several key components:

(1) \textbf{Symbolic Definition ($m_{sym}$)}. This holds the formal representation of the entity $m$ expressed in \fermat's DSL. It precisely defines the entity's logical structure. For an entity $m = \mathtt{is\_prime}$, the symbolic definition might be the following expressed programmatically,
        \[
            \begin{aligned}
            \mathit{m}_{\mathit{sym}}
              &=\;\mathtt{\lam n.}\;
                  \mathtt{\bigl(n > 1\bigr)}\;\mathtt{\land}\;
                  \mathtt{\forall\,k \in \mathbb{N}}\;.\;
                     \mathtt{\Bigl(
                        \underbrace{\mathtt{\exists\,q \in \mathbb{N}\;.\; n = q \times k}}_{\mathtt{divides(k,n)}}
                        \;\Rightarrow\;
                        (\,k = 1 \;\lor\; k = n\,)
                     \Bigr)}
            \end{aligned}
        \]

\vspace{-1em}
(2) \textbf{Computational Interpretation ($m_{comp}$)}. This is an executable Python function that provides an efficient, concrete evaluation of the entity's symbolic definition $m_{sym}$. Let $I$ be the space of potential input instances for $m$. The interpretation is a mapping $m_{comp}: I \to \{$True$,$ False, Unknown$\}$ where, for an instance $i \in I$, the function returns:

\begin{itemize}
    \item \textbf{True (}resp. \textbf{False)} if $i$ has been verified computationally (resp. not) to satisfy $m_{sym}$.
    \item \textbf{Unknown} when the evaluation of $i$ could not be determined computationally within resource limits (e.g. due to universal quantification over an infinite set).
\end{itemize}
As an example, for an entity $m=\text{square}$, its computational interpretation could be given by \linebreak $m_{comp} = \verb|lambda a, b: b == a*a|$.

(3) \textbf{Cached Instances $\mathcal{X}(m) = (\mathcal{X}^+(m), \mathcal{X}^-(m)$)}. These components store explicit input instances for the entity $m$, where $\mathcal{X}^+(m) = \{i \in I \mid m(i) = \mathrm{True}\}$ stores examples, and $\mathcal{X}^-(m) = \{i \in I \mid m(i) = \mathrm{False}\}$ stores nonexamples. 
These instances ground the entity's semantics and can be used for various purposes. For $m = \text{divides}$:
\[\mathcal{X}^+(m) = \{(2,4), (1,3), (2, 2), (3,6), \dots\}, \qquad \mathcal{X}^-(m) = \{(2, 3), (3,5), (4,1), (5,4), \dots\}\]

We write $m_i, m_o$ for input and output arity of $m$, and size$(m) = m_i + m_o$ for size of the examples.

The construction history $\mathcal{C}(m) = \{a_1, \dots, a_n\}$ of an entity $m$ is the ordered list of actions applied to produce it. Definitions ($m \in \mathcal{D}$) are further classified as either \textbf{predicates} or \textbf{functions}. This informs which production rules in the action space $\mathcal{A}$ are applicable.

\subsection{Production Rules}
Following HR \citep{colton2000hr}, \textsc{Fermat} comes equipped with a set of \emph{production rules}, consisting of composable actions for constructing new definitions and conjectures from prior ones. These rules define all construction actions in $\mathcal{A}_{def}, \mathcal{A}_{conj} \subseteq \mathcal{A}$ to produce new entities. We divide the production rules by whether they produce definitions or conjectures. We include a complete description of all the production rules present in \fermat in Appendix \ref{productionrules}, and give two condensed examples:

\textbf{Definition Production Rules.}

(1) \textbf{Exists}:  Let $\cP(\bx_1, \ldots \bx_n)$ be a predicate and $I := \{i_1, \dots, i_k\}$ be a list of indices to existentially quantify over, and let $J := \{j_1, \ldots, j_{n-k}\}$ be the remaining indices in increasing order. Then the production rule outputs a new predicate $\cQ$ as follows, 
    \[\mathtt{exists} \: \cP \: I \to \boxed{\;
    \cQ(\bx_{j_1}, \dots, \bx_{j_{n-k}}) :=  \: \exists \bx_{i_1}, \: \dots \:, \bx_{i_k} \quad \mathrm{s.t.} \quad \cP(\bx_1, \ldots, \bx_n)
    \;}\]
\textit{Example}. Consider the predicate $P(x,y)$ defined by $P(x,y) :\iff \exists z., y = x \times z$, which expresses that $x$ divides $y$. This can be constructed by applying the \textit{Exists} production rule to the predicate $\text{product}(x,z,y)$, which holds when $y = x \times z$, with the index list $I = \{1\}$ corresponding to the variable $z$ to be existentially quantified. Formally,
\[
\mathtt{exists} \: \text{product} \: I
\longrightarrow \boxed{
Q(x,y) := \exists z \: \text{s.t.} \: \text{product}(x,z,y)
}
\]

(2) \textbf{Specialize}: Given an entity, this rule outputs a new definition by specializing a variable to a fixed value. Let $\cA(\bx_1, \dots, \bx_n)$ be a function (resp. predicate), and let $i$ be the index to specialize, and $v$ be the value to substitute. Then the rule outputs a function (resp. predicate) $\cB$ as follows,
    \[\mathtt{specialize} \; \cA \; i \; v \to \boxed{\;
        \cB(\bx_1, \dots, \bx_{i-1}, \bx_{i+1}, \dots, \bx_n) := \cA(\bx_1, \dots, \bx_{i-1}, v, \bx_{i+1}, \dots, \bx_n)   
    \;}\]

\fermat contains 7 other definition production rules: (i) \emph{Compose}, which composes definitions, (ii) \emph{MapIterate}, which successively applies an iterator function, (iii) \emph{ForAll}, which universally quantifies over variables in definitions, (iv) \emph{Match}, which asserts equality of chosen variables in definitions, (v) \emph{Negate}, which outputs the negation of a concept, (vi) \emph{Size}, which outputs a definition of the cardinality of the set of inputs satisfying a condition, and (vii) \emph{Constant}, which creates constants from examples.  

\textbf{Conjecture Production Rules.} \fermat contains 4 production rules designed to construct conjectures. These are: (i) \emph{Implication}, which asserts that one definition implies another over all inputs, (ii) \emph{Equivalence}, which asserts that two definitions are equivalent, (iii) \emph{Nonexistence}, which asserts that no examples of a definition exist, (iv) \emph{Exclusivity}, which asserts that the only examples of a given definition belong to a given finite set.

\subsection{Prover}
To complete the action space, we require the ability to validate conjectures generated using \fermat's DSL. Critically, the generic DSL supports \emph{nested definitions}, facilitating modular construction of definitions and conjectures. These conjectures, which may involve previously defined concepts, are automatically constructed by our framework and passed to a backend theorem prover for verification. We instantiate this backend using the Z3 Theorem Prover \citep{Z3Tp}, and provide it with SMT-LIB input generated from our DSL via a custom-designed compiler. We choose Z3 as it is a powerful off-the-shelf black-box prover, the use of which enables us isolate the problem of synthesizing definitions and conjectures. We include examples of the Z3 support available through our DSL, and compilation down to SMT-LIB format in Appendix \ref{app:z3-info}.

\section{Learning Interestingness}
\label{seal}

\begin{wrapfigure}{r}{0.65\textwidth}
   \vspace{-0.37in}
    \centering
    \includegraphics[scale=0.35]{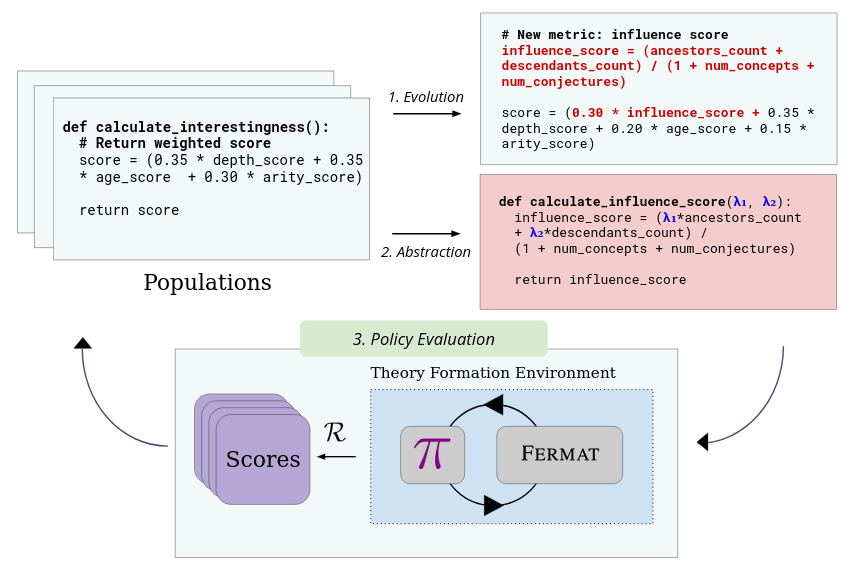}
    \caption{Overview of \method, which aims to discover an optimal interestingness measure for mathematical theory formation. It operates through three phases: (1) \textbf{Evolution}, where populations of candidate programs are generated and refined through LLM-driven mutations; (2) \textbf{Abstraction}, where high-performing programs are analyzed and reusable subroutines are extracted; and (3) \textbf{Policy Evaluation}, where the resulting programs are evaluated within the theory formation environment using \fermat, producing feedback that guides subsequent evolutionary steps.}
    \label{fig:method}
    \vspace{-1em}
    \vspace{-0.2in}
\end{wrapfigure}

In this section, we discuss our approach for automatically \emph{learning} an interestingness measure $\mathcal{I}(m, S)$ that guides the agent in discovering human mathematical knowledge. Following HR \citep{colton2000hr}, which developed simple programmatic representations of measures over features of the state, we \emph{search} in the space of Python programs that implement interestingness measures. To this end, we introduce \method (Figure~\ref{fig:method}), an evolutionary search algorithm designed to optimize an objective function given a simple numerical evaluator function. 

\paragraph{LLM-Driven Evolutionary Search.} (\textsc{EvolutionStep}). At its core, \method is an evolutionary algorithm, aimed at synthesizing programs iteratively. Each population consists of candidate interestingness programs. The generation of new programs is primarily driven by an LLM, $\mathcal{L}_{var}$, conditioned on a prompt instructing it to perform evolution. In each evolutionary step, $\mathcal{L}_{var}$ takes the program template $T$ and a selection of high-performing parent programs from a population as input, and synthesizes new candidate solutions ($f_{new}$). $\mathcal{L}_{var}$ thus acts as an operator for exploration and exploitation, intended to perform complex mutations informed by successful prior programs. We employ an island model with $k$ parallel populations ($\mathcal{P}_i$) to maintain diversity. 

\paragraph{Abstraction Learning.} (\textsc{AbstractionStep}).
Our central innovation in \method is its abstraction learning mechanism. This component is designed to identify and reuse valuable subroutines from evolved programs. This system comprises two main parts:

\begin{itemize}[leftmargin=1.2em, labelsep=0.4em, nosep]
    \item \textbf{Discovering and Utilizing Abstractions:} Periodically \method enters an abstraction phase, where an LLM $\mathcal{L}_{abs}$, analyzes a set of high-scoring programs ($S'_i$) sampled from each population. $\mathcal{L}_{abs}$ is tasked with identifying \emph{abstractions} --- valuable, reusable subroutines with defined signatures and implementations —-- within these successful programs and proposing them as new, generalized functions ($A_{new}$). These proposed abstractions are then filtered for criteria such as syntactic validity and uniqueness before being added to the island's $\text{Lib}_i$.
    
    \item \textbf{The Abstraction Library ($\text{Lib}_i$):} Each island $i$ maintains a dynamic \emph{Abstraction Library}, $\text{Lib}_i$. This library serves as a repository for functional abstractions that are identified as potentially useful during the search. Initially, these libraries are empty. After each abstraction phase, the generated abstractions are added to their respective libraries $\text{Lib}_i$. The evolutionary LLM, $\mathcal{L}_{var}$, is conditioned not only on the template $T$ and sampled programs but, crucially, also given access to the current set of abstractions in $\text{Lib}_i$ when generating new candidate programs. This encourages $\mathcal{L}_{var}$ to compose solutions by utilizing these validated sub-components, thereby promoting modularity, facilitating the construction of more complex solutions, and guiding the search towards more promising regions of the program space.
\end{itemize}

\paragraph{Policy Evaluation.} (\textsc{PolicyEvaluationStep}). In each iteration, candidate programs produced through evolution are assessed via episodic rollouts within the theory-formation environment. During a rollout, a policy instantiated by an interestingness program interacts with \fermat to guide the discovery process over multiple steps. The cumulative reward obtained across these rollouts determines each program’s fitness, providing the signal that drives subsequent evolutionary and abstraction phases.

The overall \method algorithm, detailed in Algorithm~\ref{alg:seal} (Appendix), thus proceeds in generations. Within each generation, the evolutionary search driven by $\mathcal{L}_{var}$ refines the populations on each island. Periodically, the abstraction phase mediated by $\mathcal{L}_{abs}$ enriches the abstraction libraries, which in turn provide more powerful building blocks for subsequent evolutionary steps. This interplay between LLM-driven evolution and LLM-driven abstraction learning allows \method to progressively discover and refine programmatic subroutines.

\section{Experiments}
\label{experiments}
In this section, we present empirical results evaluating the effectiveness of our approach. We aim to answer key questions about the ability of EvoAbstract to learn effective interestingness measures and the capability of \textsc{Fermat}, guided by these measures, to generate meaningful mathematical theories.

\paragraph{Environment Configuration.} \fermat centrally supports exploration in elementary number theory and finite fields, as these areas are extremely rich while easily represented. The number theory environment is supported by the Z3 Theorem Prover, while finite field reasoning is handled by a custom prover implemented in Python. For number theory, the ground truth benchmark $\mathcal{E}$ used for the extrinsic reward function $\mathcal{R}$ comprises 180 concepts, conjectures, and theorems sourced from an introductory number theory textbook \citep{andreescu2007number}, constituting a set of interesting entities. We similarly curated 67 such ground truth entities over $\mathbb{F}_{27}$, the primary finite-field setting in our experiments. These concepts span a range of theoretical sophistication, from the reflexive properties to the Goldbach conjecture. A detailed description of $\mathcal{E}$ along with subsets of the ground truth knowledge graph is detailed in Appendix \ref{app:gt}. For our experiments, we use three different starting configurations: (i) \texttt{succ\_zero\_eq} --- The definitions of zero, successor function, and the equality predicate with arity 2; (ii) \texttt{arithmetic\_base} --- Containing zero, one, two, addition, multiplication, divides, $\leq$, and the equality predicate; (iii) \texttt{ff\_27} --- Defining zero, one, and generators of $\mathbb{F}_{27}$. We include the policy template in Algorithm \ref{alg:policy_template} (Appendix).

\begin{figure}[t!]
  \centering
  
  \includegraphics[width=\linewidth]{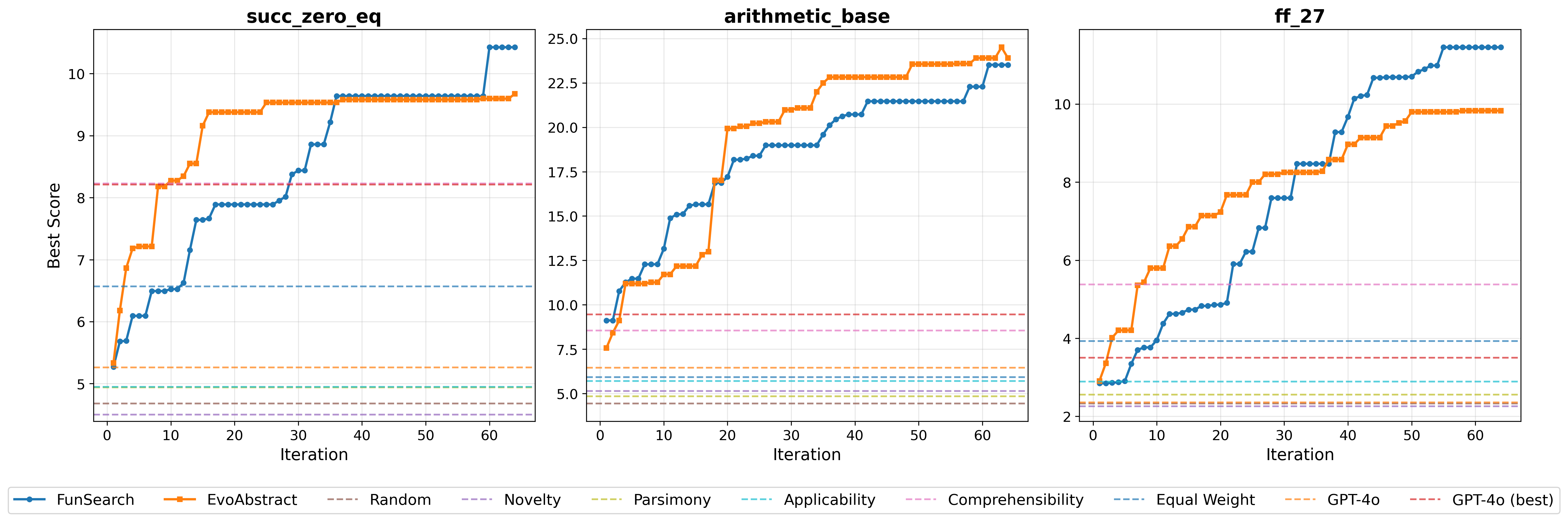}

  \caption{A plot of the best program found per iteration for FunSearch and EvoAbstract, shown for the three different starting knowledge graphs, averaged over four runs. As can be seen, the EvoAbstract and FunSearch methods dominate performance universally across all baselines. On \texttt{arithmetic\_base}, EvoAbstract slightly outperforms FunSearch, while on \texttt{succ\_zero\_eq} and \texttt{ff\_27} EvoAbstract optimizes the interestingness measures early on, but its performance plateaus sooner than FunSearch, which continues to improve.}
  \label{fig:funsearch_vs_evoabstract_multi}
\end{figure}

\paragraph{Evaluation Metrics.} To evaluate an interestingness measure, we instantiate the scoring function as extrinsic reward obtained through episodic rollouts of a policy depending on the measure through \fermat. We run $64$ episodes with a timeout of $60$ seconds\footnote{We note that we run episodes for a duration rather than a step count due to high variance in the time taken for Z3 to resolve conjectures.}, and average the reward. This evaluation metric measures the ability of the given policy to reconstruct the curriculum of human-made ground truth mathematical entities $\mathcal{E}$. We also provide qualitative analysis of the learned interestingness measures and the content of the generated theories.

\paragraph{Baselines.} We compare EvoAbstract against the following baseline methods for generating or selecting interestingness measures:
\begin{itemize}[leftmargin=1.4em, labelsep=0.4em, nosep]
    \item \textbf{Random Policy:} Selects applicable actions uniformly at random.
    \item \textbf{HR Measures:} We re-implement a number of interestingness measures manually defined in HR \citep{colton2000hr}, which operate on the state $S$ and the newly generated entity $m$ (which can be extracted from the action $a$ applied and the new state $S'$). In particular, we include the following measures:
    \begin{enumerate}
    \item \emph{Novelty}. Computes the fraction of entities with the same example classification: $M_{\text{novelty}}(m) = \# \{ m' \in S | \mathcal{X}(m) = \mathcal{X}(m')\}/\#S$.
    \item \emph{Parsimony}. Rewards a concept with fewer inputs: $M_{\text{parsimony}}(m) = \text{size}(m)^{-1}$.
    \item \emph{Productivity}. Measures how many subsequent environment steps use that entity in a production rule: $M_{\text{productivity}}(m) = \#\{m' \in S | m$ is in an action $ \in \mathcal{C}(m')\} / \# S$.
    \item \emph{Applicability}. Computes the fraction of all known instances that are examples: $M_{\text{applicability}}(m) = \mathcal{X}^+(m) / (\mathcal{X}^+(m) + \mathcal{X}^-(m))$.
    \item \emph{Comprehensibility}. Rewards a concept which is more comprehensible, measured by the inverse of the number of construction steps: $M_{\text{comprehensibility}}(m) = \# \mathcal{C}(m)^{-1}$.
\end{enumerate}
    We evaluate these measures individually and when combined in an equally weighted sum. Note that all of these measures are easily representable as Python programs.
    \item \textbf{One-shot LLM.} Instead of evolving a program, we sample $64$ programs from GPT-$4$o and evaluate their performance through episodic roll-outs, averaging the result.
    \item \textbf{FunSearch:} An ablation study where EvoAbstract is missing the abstraction component, which is equivalent to the FunSearch \citep{romera-paredes2023mathematical} algorithm without island crossover, ran at a scale afforded by our budget. We use the same hyperparameters as in our EvoAbstract evaluation.
\end{itemize}

\paragraph{EvoAbstract configuration.} We configure EvoAbstract to employ $k = 4$ islands and runs over $N_{gen} = 64$ iterations, with each interestingness function being evaluated in 16 i.i.d. rollouts. We run every configuration of EvoAbstract \& FunSearch 4 times and average the results. We instantiate both the evolution and abstraction samplers $\mathcal{L}_{var}, \mathcal{L}_{abs}$ to use GPT-$4$o-mini, and sample 2 programs per iteration. We perform the abstraction phase every 8 iterations, sampling at most two abstractions per island. $\mathcal{L}_{var}, \mathcal{L}_{abs}$ are conditioned through prompting: a system-level instruction on generating interestingness measures is attached, as well as a description of a set of Python functions which return features of the state's knowledge graph representation as well as individual entities. These functions represent the base features for the interestingness measures to manipulate. We provide a more detailed list of hyperparameters, the full prompts, and our computational resources for experiments in Appendix \ref{app:compute}.

\subsection{Experimental Results}
\label{subsec:results}

We address the following research questions:
\begin{enumerate}[labelsep=0.4em]
    \item[\textbf{RQ1:}] Can EvoAbstract effectively learn interestingness measures that outperform baseline strategies in discovering ground-truth mathematical entities?
    \item[\textbf{RQ2:}] What do the learned interestingness measures look like? Do they capture non-trivial patterns?
    \item[\textbf{RQ3:}] Can we rediscover well-known concepts in elementary number theory and finite fields?
\end{enumerate}

\paragraph{Performance Comparison (RQ1).}
We compare the cumulative extrinsic reward of policies guided by measures from \method against baselines, with results summarized in Table~\ref{fig:performance_table}. As expected, starting with a larger initial theory (\texttt{arithmetic\_base}) generally leads to greater rewards.

\begin{wrapfigure}{r}{0.65\linewidth}
  \vspace{-1em}
 \centering
 \footnotesize
 \begin{tabular}[t]{lccc}
  \toprule
  \textbf{Measure} & \makecell{\texttt{succ\_zero\_eq}}& \makecell{\texttt{arithmetic} \\ \texttt{\_base}} & \makecell{\texttt{ff\_27}} \\
  \midrule
  Random & 4.68 (2.25) & 4.44 (2.23) & 2.33 (1.20) \\
  Novelty & 4.50 (2.39) & 5.14 (2.83) & 2.26 (1.47) \\
  Parsimony & 4.94 (2.90) & 4.85 (3.09) & 2.56 (1.32) \\
  Applicability & 4.95 (2.25) & 5.71 (3.05) & 2.89 (1.69) \\
  Comprehensibility & 8.23 (2.84) & 8.55 (3.22) & 5.38 (1.89) \\
  Equal Weight & 6.57 (2.45) & 5.93 (2.82) & 3.93 (2.82) \\
  GPT‑4o & 5.26 (1.11) & 6.46 (1.98) & 2.36 (0.40) \\
  GPT-4o (best) & 8.21 (4.09) & 9.45 (3.44) & 3.50 (1.87) \\
  FunSearch & 10.23 (1.70) & 22.41 (2.68) & 11.34 (4.09) \\
  EvoAbstract & 9.62 (2.97) & 23.98 (10.50) & 9.82 (4.83) \\
  \bottomrule
 \end{tabular}
 \caption{Performance comparison of EvoAbstract against various baseline measures on three starting theories: \texttt{succ\_zero\_eq}, \texttt{arithmetic\_base}, and \texttt{ff\_27}. Each baseline receives 64 theory-formation evaluations. For FunSearch and EvoAbstract, we include the average score (standard deviation) of the best found program over four independent runs.}
 \label{fig:performance_table}
 \vspace{-2em}
\end{wrapfigure}
Among static HR measures, the random and novelty measures perform worst, exhibiting roughly equivalent scores. Parsimony's inefficacy likely stems from its limited discriminative power, as most generated definitions involve few inputs, offering insufficient signal. Comprehensibility is the strongest HR measure as it rewards simplicity of entities.
However, it alone cannot scale to more complex entities due to the combinatorial expansion of the action space.

Interestingly, the GPT-4o baseline performs only slighter better than even the random baseline (see Figure~\ref{fig:gpt-4o-measure}), and is outperformed by just the comprehensibility measure.
Despite generating more complex measures, its emphasis on rewarding construction depth and connectivity often assigns disproportionately high interestingness to initial, but irrelevant, entities. This leads to a cascading effect away from the ground truth set $\mathcal{E}$, explaining its suboptimal performance.

FunSearch \citep{romera-paredes2023mathematical} \& EvoAbstract demonstrate the value of evolutionary search.
In contrast to GPT-4o, where few generated measures surpassed the random baseline, evolutionary program synthesis yields significantly more performant measures, with the best measure discovered averaging (10.23, 22.41, 11.34) ground-truth entities per episodic roll-out on (\texttt{succ\_zero\_eq}, \texttt{arithmetic\_base}, \texttt{ff\_27}).
Incorporating the abstraction phase in \method introduces slight gains on \texttt{arithmetic\_base}, yielding measures that discover an average of 23.98 ground-truth entities, but with higher variance. Notably, on \texttt{ff\_27} and \texttt{succ\_zero\_eq}, EvoAbstract finds better solutions quicker, but the progress slows down and yields suboptimal performance at the end of the runs, on average. The abstractions are helpful in optimizing on known patterns, but produces an abstraction ``lock-in'' later on where it is difficult for the LLM to produce diverse samples that continue to increase the reward.
Beyond improved discovery, this phase also develops interpretable modular components.
Figure~\ref{fig:funsearch_vs_evoabstract_multi} illustrates the performance trajectory of EvoAbstract \& FunSearch compared to the baselines.

\paragraph{Analysis of Learned Interestingness Measures (RQ2).} 

We also conduct a qualitative analysis of the measures that EvoAbstract synthesizes.
Figures~\ref{fig:evoabstract-best-succ}, \ref{fig:evoabstract-best-ff} presents an example of the best-performing program discovered by EvoAbstract on the \texttt{succ\_zero\_eq} task.
A key characteristic of this program is its utilization of numerous abstractions and subroutines that were identified and refined during earlier abstraction phases.
These abstractions are detailed in Figures~\ref{fig:evoabstract-succ-abstractions}, \ref{fig:evoabstract-ff-abstractions}, which we now analyze.

First, EvoAbstract rediscovers and often refines variants of the baseline HR measures.
For instance, it generates applicability-like measures, such as \texttt{compute\_example\_balance}, which calculates the ratio of examples to nonexamples.
Notably, EvoAbstract can refine previous abstractions, exemplified by \texttt{calculate\_uniqueness\_score\_v2}, which generalizes prior uniqueness abstractions.
Furthermore, measures distinct from the HR baselines are found, such as \texttt{calculate\_rule\_diversity\_score}, which weighs the diversity of rules in the construction history. Additionally, it produces abstractions which generalize known construction patterns, as seen from \texttt{adjust\_score\_by\_node\_type}.

A comparison with the best program generated by FunSearch (detailed in Figure~\ref{fig:funsearch-best-succ}) is instructive.
While the FunSearch program utilizes similar components to those found by EvoAbstract, they are fewer in number. FunSearch tends to integrate these functionalities more directly, resulting in a less modular structure. The distinct modularity evident in Figure \ref{fig:evoabstract-best-succ} lends itself to quicker readability of the high-level operation of the interestingness function.

\paragraph{Analysis of Generated Theories (RQ3).} EvoAbstract \& FunSearch discover a notable portion of fundamental math concepts in our ground truth benchmark. When starting from the \texttt{succ\_zero\_eq} base, the agent successfully develops the notion of addition, multiplication, divisibility, and the tau function. Furthermore, it makes progress towards conjecturing fundamental properties of divisibility, such as the reflexivity of divisibility. When starting from \texttt{arithmetic\_base}, the agent goes further --- discovering the concepts of powers and primality along with more complex compositions of functions. In \texttt{ff\_27}, the evolutionary methods are capable of discovering concepts such as \texttt{ff\_sum\_three\_times}, but cannot find the conjecture stating the characteristic of char$(\mathbb{F}_{27}) = 3$, which requires further rule applications to discover. Relevant samples of the evolved knowledge graphs are shown in Figure \ref{fig:learnedgraphs}.

We note that the best-performing interestingness measures we find can still be suboptimal, upweighting entities are not particularly interesting to humans. For instance, we find that \texttt{equals}, which important for initial exploration, is assigned overly high interestingness, leading to an excess of redundant or vacuous statements during theory formation. While the agent generated conjectures, it had difficulty discovering many \emph{ground truth} conjectures, which is likely due to the limited correct ways to correctly specify a conjecture compared to a definition.

\subsection{Discussion}
We find that there are several avenues for further exploration towards discovering richer theories. Firstly, the policy template we employ, designed to manage combinatorial growth, exposes only a subset of complete action space at any step. This choice, while pragmatic, limits scalability to more complex mathematical objects where a lengthy list of actions must be applied in a specific order.
Secondly, we observe that there are ``bottleneck'' entities, such as primality, which must be discovered in order to continue the development of an interesting theory (see Figure \ref{fig:gt_primes}). In our experiments, when primality is discovered, the resultant knowledge graph is prohibitively large so as to obstruct valuable actions involving it.
Finally, \textsc{Fermat} does not yet exploit symmetries in entities leading to representational redundancy. For instance, while exhaustively checking for equivalences between definitions would reduce this redundancy, we found the approach to be computationally intractable with Z3 as theories grow. Further experimentation with FunSearch and EvoAbstract with heavy compute budgets will help to investigate the potential for significant discovery with evolutionary methods in these domains. Addressing these points will be crucial for advancing \textsc{Fermat}'s ability to construct deeper and more sophisticated mathematical theories.

\section{Related Works}
\textbf{Automated Theory Formation.} AM (1977) \citep{lenat1977am, lenat1983eurisko} was a theory formation program which relied on a curated set of 243 heuristics to discover concepts and prove conjectures in elementary set theory and number theory. Similarly, the Graph Theorist (1987) \citep{EpsteinGraphTheorist} performed conjecturing \& proving using an input set of definitions. HR (2000) \citep{colton2000hr} introduces a small set of production rules and manually curated heuristics to perform mathematical theory formation. Theorema (2006) \citep{BUCHBERGER2006470} performed human-in-the-loop theory exploration, leveraging computational tools in Mathematica and with an emphasis on producing human-readable proofs. Theory formation is less explored in the modern era. Notably, Minimo \citep{poesia2024learningformalmathematicsintrinsic} trains a neural model to play a game of conjecture and proof, but remains restricted to the initial axiomatic definitions. QuickSpec \citep{QuickSpec} is a symbolic theory exploration tool that interleaves term generation and random testing for conjecturing. As a final note, automated theory formation can be studied for domains other than mathematics --- BACON (1983) \citep{Langley1983} represents a program which aimed to rediscover empirical laws in chemistry.

\textbf{Conjecturing.} Many works have focused on the particular problem of synthesizing plausible conjectures. The PSLQ algorithm \citep{PSLQ} was developed for identifying integer relations between mathematical constants. Graffiti \& TxGraffiti \citep{GraffitiFajtlowicz, davila2024automatedconjecturingmathematicsemphtxgraffiti} produced conjectures in graph theory using several heuristics, given a large set of graphs and graph invariants. The Ramanujan Machine \citep{ramanujan2021} utilized several algorithms to conjecture relations between fundamental constants such as $\pi$ and $\zeta(3)$. \citet{Davies2021} uses machine learning techniques to identify patterns that lead to conjectures. 

\textbf{Theorem-Proving.} The most significant attention in modern research has been applied towards the problem of theorem-proving. Simon \& Newell's Logic Theorist and Hao Wang's Program II \citep{newell1956logic, WangMechanicalMath}, were early explorations into a theorem-proving system. Recently, neural systems \citep{dong2025stpselfplayllmtheorem, jiang2023draftsketchproveguiding, ren2025deepseekproverv2advancingformalmathematical, thakur2024incontextlearningagentformal, polu2020generativelanguagemodelingautomated} invoking interactive theorem provers like Lean \citep{de2015lean}, Isabelle \citep{paulson1994isabelle}, and Coq \citep{Coq} have seen great interest. AlphaProof \citep{deepmind2024alphaproof} and AlphaGeometry \citep{trinh2024solving} together attained a silver medal at the International Mathematical Olympiad. Another interesting angle for resolving conjectures comes from use of SAT \& SMT solvers --- notably, yielding a resolution to the Boolean Pythagorean Triples problem \citep{Heule_2016}. Notably, the Four Colour Theorem was proved through computer-assisted case-checking \citep{ROBERTSON19972}. Similarly, \citep{ charton2024patternboostconstructionsmathematicslittle} used a combination of neural and symbolic techniques to disprove conjectures.

\textbf{Program Synthesis \& RL.} FunSearch \citep{romera-paredes2023mathematical} and  AlphaEvolve\citep{AlphaEvolve} are LLM-guided evolutionary algorithms used to discover programs producing mathematical constructions. The closely related LaSR \citep{grayeli2024symbolicregressionlearnedconcept} algorithm uses LLM-guided evolution and a learned \emph{textual} abstraction library for symbolic regression. Eureka \citep{ma2024eurekahumanlevelrewarddesign} uses iterative LLM refinement to produce extrinsic reward functions that outperform human-engineered rewards on a suite of RL environments. Several efforts \citep{Bowers_2023, ellis2020dreamcodergrowinggeneralizableinterpretable, grand2024lilolearninginterpretablelibraries} perform program synthesis in functional languages and develop abstraction libraries using symbolic abstraction algorithms.  An interesting direction would be to develop separate explorative and exploitative policies, as in \citep{norman2024firstexploreexploitmetalearningsolve}, for mathematical theory formation.

\section{Conclusion}
In this work, we targeted the problem of capturing the interestingness of concepts in the context of open-ended mathematical discovery. To support our research, we introduced \fermat, a novel RL environment for theory formation. We introduced \method, an LLM-based evolutionary procedure which abstracts and stores useful subroutines identified during search. We show that our learned interestingness measures outperforms several baselines when conducting theory formation using \fermat, starting from basic definitions, in elementary number theory and finite fields.

Our investigation is a starting point for much broader research in automated theory formation. Integrating interactive theorem provers, like Lean \citep{de2015lean}, into \fermat will allow exploration in more complex domains and enable studying the problem of \emph{learning to prove tabula rasa}. It is also an open problem how to autonomously synthesize production rules.  In future work, we see that extensions of \fermat could lead to the development of new mathematics as envisioned in the early days of Artificial Intelligence.

\section{Acknowledgement}
This research was supported in part by NSF awards CCF-2212559 and CCF-2403211, a grant from Renaissance Philanthropy's AI for Math Fund, and the NSF AI Institute for Foundations of Machine Learning. We would also like to thank the anonymous reviewers for their insightful feedback, which helped improve the quality of this manuscript.

\bibliographystyle{neurips_conference}
\bibliography{neurips_conference}

\newpage

\appendix

\section{Appendix / supplemental material}
\subsection{Production Rules.}
\label{productionrules}

Here we include a full description of each production rule available in \fermat, including those mentioned in the main content for completeness.

\paragraph{Definition Production Rules.}

\begin{enumerate}
    \item \textbf{Compose}: This rule allows for composition of functions, predicates, and function-to-predicates. 
    
        \forceindent \textbf{Function Composition}. Let 
            \[\cF: X_1 \times \ldots \times X_n \to Y_1 \times \ldots \times Y_m, 
            \qquad \cG: Z_1 \times \ldots \times Z_k \to W\] 
            be functions, and $I: \{1, \dots, m\} \to \{1, \dots, k\}$ be a map from $\cF$'s output indices to $\cG$'s input indices. Let $\bx_1, \ldots, \bx_n$ denote the parameters to be passed into $\cF$, and $\bp_1, \ldots, \bp_i$ denote any additional parameters to be passed to $\cG$. Then the production rule outputs a function as follows,
            \[\mathtt{compose} \: \cF \: \cG \: I \: \to \: \boxed{\;\cH(\bx_1, \ldots, \bx_n, \bp_1, \ldots, \bp_i) := \cG(\bv_1, \ldots, \bv_k)\;}\] where
                \[\bv_{j} \;
            \begin{cases}
              = \cF(\bx_{1},\dots,\bx_{n})_{i}, & \text{if } j = I(i), \\[6pt]
              \in \{\bp_1, \ldots, \bp_{i}\},                    & \text{if } j \notin \mathrm{Image}(I).
            \end{cases}
        \]
        \forceindent \textbf{Predicate Composition}. Let
        \[
          \cP : X_{1} \times \dots \times X_{n} \;\longrightarrow\; \text{Bool},
          \qquad
          \cQ : Z_{1} \times \dots \times Z_{k} \;\longrightarrow\; \text{Bool}.
        \]
        be predicates, and $S: \{1, \dots, n\} \to \{1, \dots, k\}$ be a sharing map from $\cP$'s input variables to $\cQ$'s input variables, that is, $S(i) = j$ if the $i$-th input variable of $\cP$ and $j$-th input variable of $\cQ$ will be shared when constructing the output predicate $\cR$ defined below. Let $\mathrm{Image}(S) = \{i_1, \ldots, i_s\}$ with $i_1 < \ldots < i_s$. Then the production rule outputs a predicate as follows,
        \[\mathtt{compose} \: \cP \: \cQ \: S \: \to  \boxed{\;
          \cR(\bx_{1},\dots,\bx_{n},\,\bp_{1},\dots,\bp_{i})
          :=
          \cP(\bx_{1},\dots,\bx_{n}) \;\land\; \cQ\bigl(\bv_{1},\dots,\bv_{k}\bigr)
          \;}
        \]
        where
        \[\bv_{j} \;\;
            \begin{cases}
              = \bx_i, & \text{if } j = I(i), \\[6pt]
              \in \{\bp_1, \ldots ,\bp_{i}\},                    & \text{if } j \notin \mathrm{Image}(I).
            \end{cases}
        \]
        \forceindent \textbf{Function to Predicate Composition}. This case works identically to function composition. Let 
            \[\cF: X_1 \times \ldots \times X_n \to Y_1 \times \ldots \times Y_m, 
            \qquad \cP: Z_1 \times \ldots \times Z_k \to \mathrm{Bool}\] 
            be a function and a predicate, and $I: \{1, \dots, m\} \to \{1, \dots, k\}$ be a map from $\cF$'s output indices to $\cP$'s input indices. Let $\bx_1, \ldots, \bx_n$ denote the parameters to be passed into $\cF$, and $\bp_1, \ldots, \bp_i$ denote any additional parameters to be passed to $\cP$. Then the production rule outputs a predicate as follows,
            \[\mathtt{compose} \: \cF \: \cP \: I \: \to \:\boxed{\;\cH(\bx_1, \ldots, \bx_n, \bp_1, \ldots, \bp_i) := \cP(\bv_1, \ldots, \bv_k)\;}\] where
                \[\bv_{j} \;
            \begin{cases}
              = \cF(\bx_{1},\dots,\bx_{n})_{i}, & \text{if } j = I(i), \\[6pt]
              \in \{\bp_1, \ldots, \bp_{i}\},                    & \text{if } j \notin \mathrm{Image}(I).
            \end{cases}
        \]
    \item \textbf{Exists}: This rule allows for existentially quantifying out variables in a predicate or function. 
    
    \forceindent\textbf{Predicate}. Let $\cP(\bx_1, \ldots \bx_n)$ be a predicate, and let $I := \{i_1, \dots, i_k\}$, where $k < n$, be a list of input indices to existentially quantify over. Let $J := \{1, \dots, n\} \setminus I = \{j_1, \ldots, j_{n-k}\}$ with $j_1 < \dots j_{n-k}$ be the remaining indices. Then the production rule outputs a new predicate $\cQ$ as follows, 
    \[\mathtt{exists} \: \cP \: I \to \boxed{\;
    \cQ(\bx_{j_1}, \dots, \bx_{j_{n-k}}) :=  \: \exists \bx_{i_1}, \: \dots \:, \bx_{i_k} \quad \mathrm{s.t.} \quad \cP(\bx_1, \ldots, \bx_n)
    \;}\]
    \forceindent \textbf{Function}. Let $\cF(\bx_1, \ldots \bx_n)$ be a function, and let $I$ and $J$ be defined similarly as before. Then the production rules outputs a new predicate $\cQ$ as follows,
    \[\mathtt{exists} \: \cF \: I \to \boxed{\;
    \cQ(\bx_{j_1}, \dots, \bx_{j_{n-k}}, \: \by) :=  \: \exists \bx_{i_1}, \: \dots \:, \bx_{i_k} \quad \mathrm{s.t.} \quad \cF(\bx_1, \ldots, \bx_n) = \by.
    \;}\]
    \item \textbf{Map Iterate}: This rule turns an iterator function (unary or binary) into a new function by applying the iterator function $n$ times.  

    \forceindent \textbf{Unary Function}. Let $\cF$ be a unary iterator function. Let $\cF^n$ be the $n$-fold application of $\cF$, that is, $\cF^n(x) = \underbrace{\cF\!\bigl(\cF\!\bigl(\dots \cF}_{n\ \text{times}}\!(x)\bigr)\dots\bigr)$. Then this rule outputs a new function $\cG$ as follows,
    \[\mathtt{map\_iterate} \: \cF \to \boxed{\;\cG(\bx, n) := \cF^n(\bx)\;}.\]

    \forceindent \textbf{Binary Function}. Let $\cF$ be a binary iterator function, and $v$ be an initial value concept to be passed into the iterator. Then this production rule outputs a new function $\cG$ as follows,
      \[\mathtt{map\_iterate} \: \cF \: v \to \boxed{\;
          \begin{aligned}
              \cG(\bx,0)   &:= v,\\
              \cG(\bx,n+1) &:= \cF(\cG(\bx,n),\,\bx).
          \end{aligned}
      \;}
      \]
    
    \item \textbf{Forall}: This rule allows for universal quantification of variables over either one or two predicates. 
    
    \forceindent \textbf{Single Predicate}. Let $\cP(\bx_1, \dots, \bx_n)$ be a predicate, and let $U = \{u_1, \dots, u_j\}$ be a list of indices of the variables to universally quantify over. Let $\bar{U} = \{\bar{u}_1, \dots, \bar{u}_{n-j}\}$ be the remaining indices of the free variables. The production rule outputs a new predicate $\cR$ such that 
    \[\mathtt{forall} \; \cP \; U\to \boxed{\;
    \cR(\bx_{\bar{u}_1}, \dots, \bx_{\bar{u}_{n-j}}) := \qquad \forall \bx_{u_1}, \dots, \bx_{u_j}, \quad \cP(\bx_1, \dots, \bx_n)
    \;}.\]
    
    \forceindent \textbf{Two Predicates}. Let $\cP(\bx_1, \dots, \bx_n)$ and $\cQ(\by_1, \dots, \by_k)$ be predicates. Let $S \subseteq \{1, \dots, n\} \times \{1, \dots, k\}$ be a one-to-one sharing map, so that whenever $(i, j) \in S$, we identify the variables $\bx_i$ and  $\by_j$ by substituting $\by_j$ with $\bx_i$.  
    
    Define the merged variable set $\mathbf{M} := \{\bm_1, \dots, \bm_{n + k - |S|}\}$ where the first $n$ variables are $\bx_1, \dots, \bx_n$ in order, and the next $k - |S|$ variables are $\by_i$ variables where $i$ does not appear in the second component of any pair in $S$, indexed in ascending order. Let $\bm_{\tau(i)}$ denote the variable in $\mathbf{M}$ corresponding to $\by_i$. 
    
    Define the universal quantifier set $U \subseteq \{1, \dots, n + k - |S|\}$ to be the set of indices of variables in $\mathbf{M}$ to quantify over. Let $\bar{U}$ denote the remaining indices of the free variables. Letting $U = \{u_1, \dots, u_j\}$ and $\bar{U} = \{\bar{u}_1, \dots, \bar{u}_{n + k - |S| - j}\}$, the production rule outputs a new predicate $\cR$ such that
    \[\mathtt{forall}\; \cP \; \cQ \; S \; U\ \to\]
    \[\boxed{\;
    \cR(\bm_{\bar{u}_1}, \dots, \bm_{\bar{u}_{|\bar{U}|}}) := 
    \forall \bm_{u_1}, \dots, \bm_{u_j}, \; 
    \cP(\bm_1, \dots, \bm_n) \implies \cQ(\bm_{\tau(1)}, \dots, \bm_{\tau(k)}).
    \;}\]

    \item \textbf{Match}: This rule allows variables to be set equal to each other. Let $\cA(\bx_1, \dots, \bx_n)$ be a function (resp. predicate), and let $I := \{i_1, \dots, i_k\}$ with $i_1 < \dots i_k$ be a set of indices to be matched. Let $J := (\{1, \dots, n\} \setminus I) \cup \{i_1\} = \{j_1, \ldots, j_{n-k+1}\}$ with $j_1 < \dots j_{n-k+1}$. Then the rule outputs a new function (resp. predicate) $\cB$ with $n - k + 1$ arguments satisfying 
\[\mathtt{match} \; \cA \; I \to
  \boxed{\;
    \cB\!\bigl(\bx_{j_{1}},\dots,\bx_{j_{\,n-k+1}}\bigr)
    \;:=\;
    \cA(\bx_{1},\dots,\bx_{n})
    \;\Bigl|_{\;\bx_{i_{2}},\dots,\bx_{i_{k}}\;\gets\;\bx_{i_{1}}}
  \;}
\]

—that is, every occurrence of $\bx_{i_{2}},\dots,\bx_{i_{k}}$ in $\cA$ is replaced by the variable $\bx_{i_{1}}$.
    \item \textbf{Constant}: This rule can turn an example into a concept (that accepts no inputs, i.e. a value concept). In \fermat, we make the distinction between concepts and examples, and this rule provides a convenient way to bridge the gap.  Let $e \in \mathcal{X}^+(m)$ be an example of the concept $m$. Then the production rule synthesizes a value concept $E$ out of this example, 
    \[\mathtt{constant} \; e \to \boxed{\;
        E
    \;}\]
    \item \textbf{Specialize}: This rule allows for specializing the input (for functions and  predicates) or output (for functions). 

    \forceindent \textbf{Specialize Input.} Let $\cA(\bx_1, \dots, \bx_n)$ be a function (resp. predicate), and let $i$ be the index to specialize, and $v$ be the value to substitute. Then the rule outputs a new function (resp. predicate) $\cB$ satisfying 
    \[\mathtt{specialize} \; \cA \; i \; v \to \boxed{\;
        \cB(\bx_1, \dots, \bx_{i-1}, \bx_{i+1}, \dots, \bx_n) := \cA(\bx_1, \dots, \bx_{i-1}, v, \bx_{i+1}, \dots, \bx_n)   
    \;}\]

    \forceindent \textbf{Specialize Output.} Let $\cF(\bx_1, \dots, \bx_n)$ be a function and let $v$ be the value we want to specialize the concept to. Then the rule outputs a new predicate $\cP$ satisfying
    \[\mathtt{specialize} \; \cF \; v \to \boxed{\;
        \cP(\bx_1, \dots, \bx_n) :=  (\cF(\bx_1, \dots, \bx_n) = v)
    \;}\]

    \item \textbf{Negate}: Let $\cP$ be a predicate, then this production rule outputs the negation of the predicate, i.e. \[\mathtt{negate} \; \cP \to \boxed{\;\cQ := \mathrm{Not}(\cP)\;}\]
    \item \textbf{Size}: Let $\cP(\bx_1, \dots, \bx_n)$ be a predicate and $I = \{i_1, \dots, i_m\}$ be a set of indices. Let $J := \{1, \dots, n\} \setminus I = \{j_1, \ldots, j_{n-m}\}$ with $j_1 < \dots j_{n-k}$. Then the production rule outputs a new concept $\cQ$ 
    \[\mathtt{size} \; \cP \; I \to \boxed{\;
    \cQ(\bx_{j_1}, \dots, \bx_{j_{n-m}}) := \#\{ (\bx_{i_1}, \dots , \bx_{i_m}) \,|\, \cP(\bx_1, \ldots, \bx_n)\}
    \;}\]
    where $\#X$ denotes the cardinality of the set $X$. 
\end{enumerate}

\paragraph{Conjecture Production Rules.}
\begin{enumerate}
    \item \textbf{Implication}: Let $\cP$ and $\cQ$ be predicates over the same domain, each with $n$ inputs. Then the production rule outputs a conjecture 
    \[\mathtt{implies} \; \cP \; \cQ \to \boxed{\;\forall \bx_1, \dots, \bx_n, \cP(\bx_1, \dots, \bx_n) \implies \cQ(\bx_1, \dots, \bx_n)\;}\]
    \item \textbf{Equivalence}: This rule conjectures equivalence of concepts. 

    \forceindent \textbf{Predicate.} Let $\cP$ and $\cQ$ be predicates over the same domain, each with $n$ inputs. Then the production rule outputs a conjecture 
     \[\mathtt{equivalence} \; \cP \; \cQ \to\boxed{\;\forall \bx_1, \dots, \bx_n, \cP(\bx_1, \dots, \bx_n) \iff \cQ(\bx_1, \dots, \bx_n)\;}\]

    \forceindent \textbf{Function.} Let $\cF$ and $\cG$ be functions over the same domain, each with $n$ inputs. Then the production rule outputs a conjecture 
    \[\mathtt{equivalence} \; \cF \; \cG \to\boxed{\;\forall \bx_1, \dots, \bx_n, \cF(\bx_1, \dots, \bx_n) = \cG(\bx_1, \dots, \bx_n)\;}\]

    \item \textbf{Nonexistence}: This rule asserts non-existence conjectures.

    \forceindent \textbf{Predicate.} Let $\cP(\bx_1, \dots, \bx_n)$ be a predicate. Then the production rule outputs a conjecture 
    \[\mathtt{nonexistence} \; \cP \to \boxed{\;
        \not\exists \bx_1, \dots, \bx_n, \; \cP(\bx_1, \dots, \bx_n)
    \;}\]

    \forceindent \textbf{Function.} Let $\cF(\bx_1, \dots, \bx_n)$ be a function and $v$ be a value. Then the production rule outputs a conjecture 
    \[\mathtt{nonexistence} \; \cF \; v \to\boxed{\;
        \not\exists \bx_1, \dots, \bx_n, \; \cF(\bx_1, \dots, \bx_n) = v
    \;}\]

    \item \textbf{Exclusivity}: This production rule outputs conjectures stating that certain concepts are satisfied only on a particular finite set of inputs. 

    \forceindent \textbf{Predicate.} Let $\cP(\bx_1, \dots, \bx_n)$ be a predicate, and $S$ be a subset of $\mathrm{Domain}(\cP)$. Then the production rule outputs a conjecture 
    \[\mathtt{exclusivity} \; \cP \; S \to\boxed{\;
        \forall \bx_1, \dots, \bx_n, \; \cP(\bx_1, \dots, \bx_n) \implies (\bx_1, \dots, \bx_n) \in S
    \;}\]

    \forceindent \textbf{Function.} Let $\cF(\bx_1, \dots, \bx_n)$ be a function and $v$ be a value. Then the production rule outputs a conjecture 
    \[\mathtt{exclusivity} \; \cF \; S \; v \to \boxed{\;
        \forall \bx_1, \dots, \bx_n, \; \cF(\bx_1, \dots, \bx_n) = v \implies (\bx_1, \dots, \bx_n) \in S
    \;}\]
    
\end{enumerate}

A production rule application will propagate the input entities' computational implementation and examples where possible. While the produced symbolic definitions and computational implementations are deterministic, some rules have nondeterminism in the manner which new examples are adding upon creating.

\subsection{\textsc{Fermat} Technical Details.}
Here we describe further implementation details regarding \fermat. 
\begin{enumerate}
    \item \textbf{Forbidden paths}: Following HR, we institute some forbidden paths which disallow the application of certain rules automatically. In particular, this disallows the creation of the following definitions \& conjectures on input definition $P$: $[\lnot \lnot P, P \implies P, P \iff P, P \iff \lnot P]$. Though this forms a minor optimization for our experiments, we believe that this set of forbidden paths can be learned automatically. In particular, in an extension of \fermat which also allows interpretable proofs, an agent may quickly recognize these paths lead to uninteresting proofs and prevent their usage in the future.
    \item \textbf{Global Instance Storage}: By instance of a theory, we refer to all concrete values of the domain introduced thus far in the theory. Our environment keeps track of all instances seen in the theory throughout the theory exploration process.
    \item \textbf{Z3 Example Addition}: Definitions created involving the universal or existential quantifier rules cannot add examples or non-examples respectively. This is because adding such instances to the data of the entity requires iterating over all values of the Nat type, which is infinite. However, the more data an entity has the richer the theory. For such cases, we add certified examples for such definitions, by randomly sampling an instance of values, and using Z3 to determine whether it forms an example or non-example. We find this helps to prevent future nonexistence and trivial implication conjectures.
\end{enumerate}

\subsection{Proving through Z3.}
\label{app:z3-info}

Our DSL allows users to define functions and predicates over bounded and unbounded parameters, and to compose them into logical conjectures using constructs such as \texttt{ForAll}, \texttt{Exists}, \texttt{Implies}, \texttt{And}, and arithmetic expressions. Critically, the DSL supports \emph{nested definitions}: a predicate can define helper functions and other predicates inside its body, and similarly for functions. This enables the modular construction of conjectures and facilitates reuse of previously discovered building blocks.

\paragraph{Compiler and SMT-LIB Translation.}  
The DSL is compiled to SMT-LIB, the input language accepted by Z3. Our compiler performs:
\begin{enumerate}
    \item Flattening of nested definitions into top-level SMT functions,
    \item Lexical scoping resolution and name hygiene to avoid collisions,
    \item Translation of DSL-level constructs into logically equivalent SMT forms. We write a compiler which converts our DSL into the SMT-lib target language. The compiler is written using the \texttt{parglare} \citep{dejanovic2021b} library. 
\end{enumerate}

\begin{figure}[ht]
  \centering
  \begin{subfigure}[t]{0.48\textwidth}
    \lstset{language=Python,basicstyle=\ttfamily\small,frame=single, framesep=5pt}
    \begin{lstlisting}
    f_0 := Func(
      params 1;
      bounded params 0;
      ReturnExpr 2 * x_0;
      ReturnPred None;
    );
    f_1 := Func(
      params 0;
      bounded params 0;
      ReturnExpr 6;
      ReturnPred None;
    );
    ReturnExpr None;
    ReturnPred Exists(
        [b_0], 
        f_0(x_0=b_0) == f_1()
    );
    \end{lstlisting}
    \caption{A DSL program asserting that 6 is even using function composition.}
    \label{fig:even_dsl}
  \end{subfigure}
  \hfill

  \begin{subfigure}[t]{0.48\textwidth}
    \lstset{language=Python,basicstyle=\ttfamily\small,frame=single,framesep=5pt}
    \begin{lstlisting}
    p_0 := Pred(
      params 1;
      bounded params 1;
    
      f_0 := Func(
        params 1;
        bounded params 0;
        ReturnExpr x_0 + 1;
        ReturnPred None;
      );
    
      ReturnExpr None;
      ReturnPred Exists(
        [b_0], 
        f_0(x_0=b_0) == x_0
      );
    );
    \end{lstlisting}
    \caption{A nested DSL predicate defining an inner function and using it in an existential condition.}
    \label{fig:nested_dsl}
  \end{subfigure}
  
  \caption{DSL snippets}
  \label{fig:dsl_code_figure}
\end{figure}

\begin{figure}[h]
    \centering
    \lstset{language=Lisp,basicstyle=\ttfamily\small,frame=single}
    \begin{lstlisting}
(define-fun f_0_p_0 ((x_0 Int)) Int (+ x_0 1))
(define-fun p_0 ((x_0 Int)) Bool
  (exists ((b_0 Int)) (= (f_0_p_0 b_0) x_0)))
    \end{lstlisting}
    \caption{SMT-LIB code generated from Figure~\ref{fig:nested_dsl}, where naming collisions are avoided by hygienic flattening.}
    \label{fig:smtlib_translation}
\end{figure}

\paragraph{Semantics and Proof Feedback.}  
When a conjecture is compiled and passed to Z3, the prover returns one of three outcomes:

\begin{description}
    \item[\texttt{UNSAT}] — The conjecture is logically valid. It is added to the theory as a verified theorem and can be used in future derivations.
    \item[\texttt{SAT}] — The conjecture is invalid. Z3 returns a counterexample, which is parsed back into the DSL domain.
    \item[\texttt{Timeout}] — We used 2s timeout with the Z3 solver. 
\end{description}

\paragraph{Theory-Guided Composition.}  
The DSL serves as a unifying language in our system: all definitions, lemmas, and conjectures are expressed as DSL programs. As new concepts are discovered by our theory formation framework (\fermat), they are registered as definitions in the DSL. New conjectures are then automatically generated by composing these building blocks. This compositional ability—enabled by nesting and a hygienic compiler—allows our system to express and verify arbitrarily structured mathematical ideas.

\subsection{Ground Truth Set}
\label{app:gt}

\begin{figure}[ht]
\centering
\includegraphics[width=0.8\textwidth]{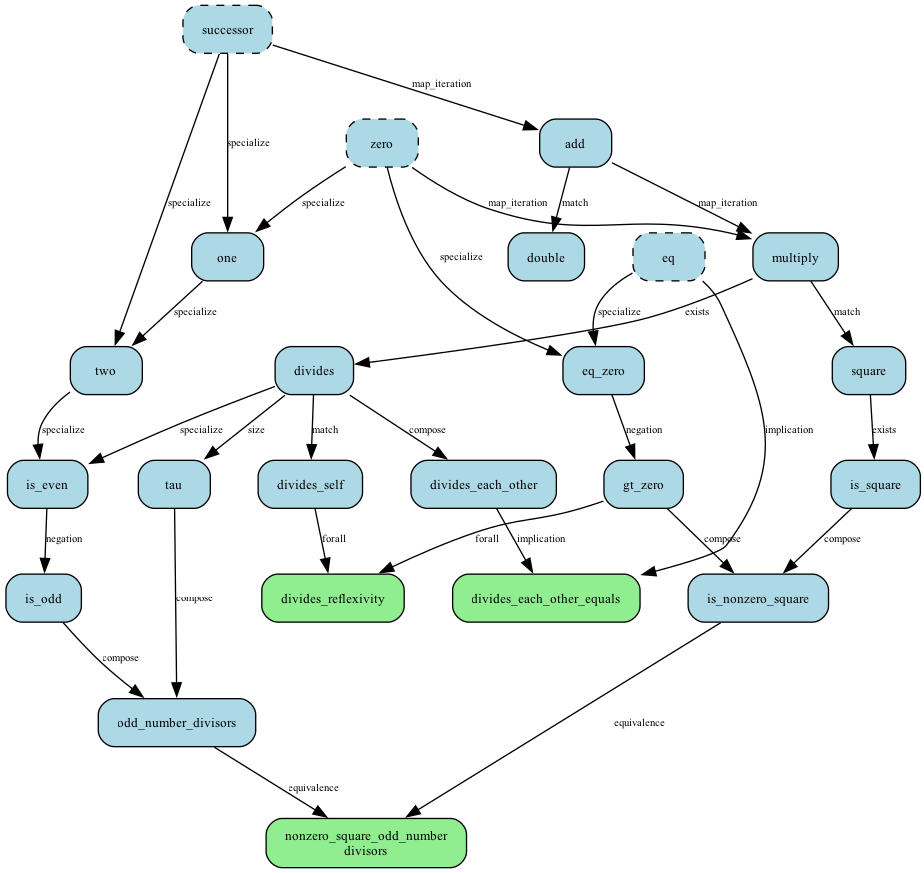} 
\caption{Sample knowledge graph of ground truth entities relating to basic properties of divisibility in the domain of natural numbers.}
\label{fig:gt_basics}
\end{figure}

\begin{figure}[ht]
\centering
\includegraphics[width=0.9\textwidth]{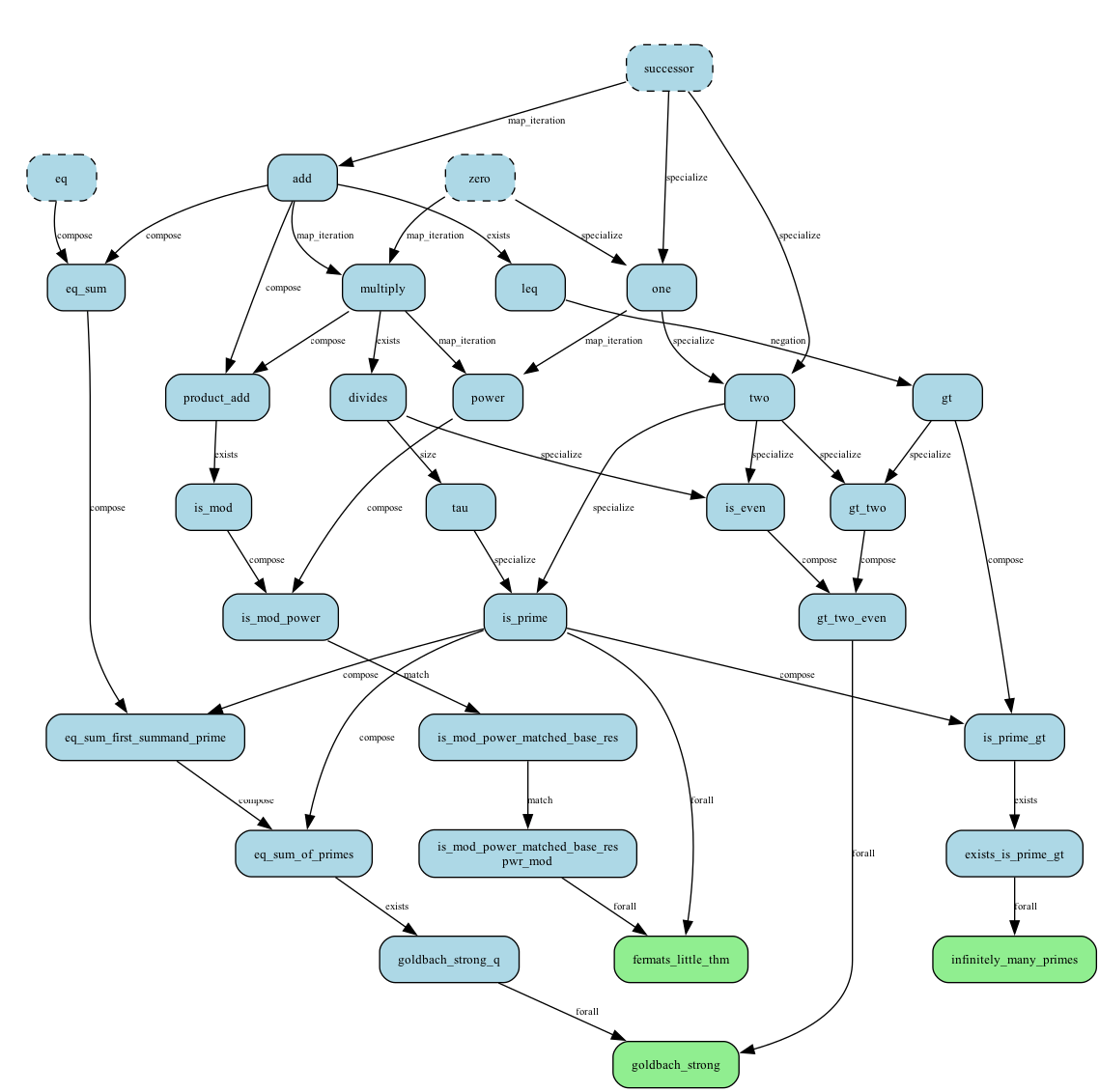} 
\caption{Sample knowledge graph of ground truth entities relating to theorems and conjectures central to the theory of prime numbers. Note that the concept of primality, \texttt{is\_prime}, is an ancestor of many concepts.}
\label{fig:gt_primes}
\end{figure}

\begin{figure}[p]
    \centering
    \caption{Sample knowledge graph of ground truth entities relating to theorems and conjectures central to the theory of the finite field $\mathbb{F}_{27}$.}

    \rotatebox{90}{\includegraphics[height=\textwidth/2]{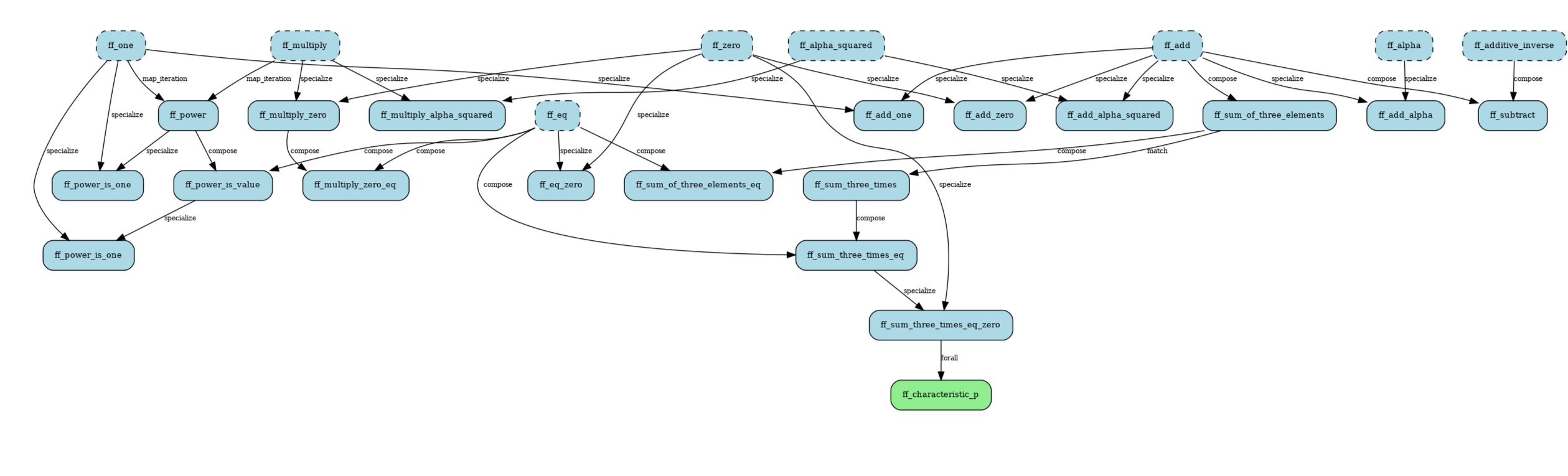}}
    \label{fig:gt_ff}
\end{figure}

We curated 180 ground truth functions, theorems, and conjectures using a well-known introductory elementary number theory text \cite{andreescu2007number} as well as a small set of famous conjectures in the number theory literature, and an additional 67 ground truth entities drawn from the theory of finite fields over $\mathbb{F}_{27}$. These ground truth concepts can be entirely derived by applying the production rules to the base concepts ( \texttt{zero}, \texttt{successor}, and \texttt{equality} for number theory, and generators and field operations for $\mathbb{F}_{27}$). Figure~\ref{fig:gt_basics} illustrates a small subset of ground truth concepts that relate to divisibility, demonstrating an area in which the model would receive extrinsic reward for discovering the most basic properties of natural numbers. On the other hand, Figure~\ref{fig:gt_primes} illustrates a small subset of ground truth concepts relating to more sophisticated theorems and conjectures in the theory of prime numbers which offer extrinsic reward in theorizing about abstract ideas such as the existence of infinite instances. Figure~\ref{fig:gt_ff} covers a subset of definitions and theorems in $\mathbb{F}_{27}$.

It is worth noting that there may be multiple paths to arriving at a single concept. For example, it is possible to derive the concept \texttt{is even} either by applying:

\begin{verbatim}
    apply specialized divides two index_to_specialize=0
\end{verbatim}

which gives a new concept with the first argument of the \texttt{divides} function to two, or by applying:

\begin{verbatim}
    apply exists double indices_to_quantify=0
\end{verbatim}

which gives a new concept that returns \texttt{True} for an input if there exists a natural number such that doubling it results in the input. The number of possible paths to reach a certain ground truth concept increases exponentially with complexity. Because we want to evaluate the algorithm on its ability to find a ground truth concept regardless of the path it takes, we included redundant concepts in our ground truth set. In this manner, we cover as many paths to meaningful math concepts as possible, and we provide a smooth reward signal to the algorithm for defining interestingness.

\subsection{Computational Resources \& Hyperparameters}
\label{app:compute}
Our experiments are run on 64 Intel Xeon Platinum 8352Y and 64 AMD EPYC 7413 24C CPUs. We leverage parallelism built into \textsc{Fermat} to enable speedup in the evaluations. Given this allocation, evaluating a single interestingness measure through episodic rollouts with \textsc{Fermat} using the configuration detailed in \label{experiments} takes $\sim M = 120$ seconds when using $64$ workers. Our FunSearch and EvoAbstract experiments take significantly longer due to large number of interestingness measures generated and evaluated during evolutionary search. In particular, each experimental result reported with either FunSearch/EvoAbstract takes approximately $18$ hours with $64$ workers. The evaluation of our GPT-$4$o baseline takes a total of $6$ hours when using $64$ workers.

\vspace{-1pt}
\begin{figure}[H]
    \centering
    \begin{algorithm}[H]
        \caption{EvoAbstract: Synthesis via Evolution and Abstraction Learning}
    \label{alg:seal}
\algtext*{EndFor}
\algtext*{EndIf}      
\algtext*{EndWhile} 
\begin{algorithmic}[1] 
\Require Template $T$; Number of islands $k$; Generations $N_{gen}$; Abstraction frequency $G \in \mathbb{N}^+$; 
\Require Parent sample size $n_p \in \mathbb{N}^+$; Abstraction candidate sample size $n_{abs} \in \mathbb{N}^+$; Evolution sample size $n_e$; 
\Require Evolution \& Abstraction LLMs $\mathcal{L}_{var}, \mathcal{L}_{abs}$;
\State Initialize $k$ populations $\mathcal{P}_1, \dots, \mathcal{P}_k$ with seed programs.
\State Initialize $k$ empty abstraction libraries $\text{Lib}_1, \dots, \text{Lib}_k$.

\For{generation $g = 1$ to $N_{gen}$}
    \State Sample island $i \sim \text{Uniform}\{1, \dots, k\}$.
    \State $P \leftarrow$\textsc{EvolutionStep}$(\mathcal{P}_i, T, \text{Lib}_i, n_p, n_e, \mathcal{L}_{var}, \mathrm{Scores})$.
    \State $\mathrm{Score} \leftarrow \textsc{PolicyEvaluationStep}(P)$.
    \State Update population $\mathcal{P}_i \gets \mathcal{P}_i \,\cup\, \{P\}$
    \If{$g \bmod G \equiv 0$} \Comment{Perform abstraction phase periodically}
        \ForAll{islands $i = 1$ to $k$}
            \State $\text{Lib}_i \gets \text{Lib}_i \,\cup\, $ \textsc{AbstractionStep}$(\mathcal{P}_i, T, \text{Lib}_i, n_{abs}, \mathcal{L}_{abs}, \mathrm{Scores})$.
        \EndFor
    \EndIf
\EndFor
\State \Return Best program $f^*$ found across all populations $\mathcal{P}_1, \dots, \mathcal{P}_k$.
\end{algorithmic}
\end{algorithm}

    \label{fig:method-pseudocode}
    \vspace{-1.8em}
\end{figure}

\begin{algorithm}[H] 
\caption{Policy Template for Action Selection}
\label{alg:policy_template}
\begin{algorithmic}[1] 

\Require Interestingness measure $\mathcal{I}(\text{entity}, \text{graph}) \mapsto \mathbb{R}$.
\Require Knowledge Graph $G = (V,E)$ with definitions $\mathcal{D} \subset V$.
\Require Number of definitions to sample $N \in \mathbb{N}^+$.
\Require Simulation limit $S_{lim} \in \mathbb{N}^+$.

\State $\mathcal{D}_{sampled} \gets \emptyset$ \Comment{Set of $N$ sampled definitions}
\State $Scores \gets \{\}$ \Comment{Map each definition to its interestingness score}
\For{each definition $d \in \mathcal{D}$}
    \State $Scores[d] \gets \mathcal{I}(d, G)$
\EndFor
\State $\mathcal{D}_{sampled} \gets \text{SampleByScore}(Scores, N)$

\State $\mathcal{A}_{potential} \gets \text{EnumeratePossibleActions}(\mathcal{D}_{sampled}, G)$

\State $\mathcal{A}_{sim} \gets \text{Sample}(\mathcal{A}_{potential}, \min(S_{lim}, |\mathcal{A}_{potential}|))$
    \Comment{Randomly sample up to $S_{lim}$ actions for simulation}

\State $SimulatedActionScores \gets \{\}$
\For{each action $a \in \mathcal{A}_{sim}$}
    \State $e_{new} \gets \text{SimulateAction}(a, G)$ \Comment{Simulate action $a$ to get resulting entity $e_{new}$}
    \State $score_a \gets \mathcal{I}(e_{new}, G)$ \Comment{Compute interestingness of the new entity}
    \State $SimulatedActionScores[a] \gets score_a$
\EndFor

\State $a^* \gets \text{SampleByScore}(\mathcal{A}_{sim}, 1)$

\State \Return $a^*$
\end{algorithmic}
\end{algorithm}

We note that our episodic roll-outs are \emph{time-capped}, not finishing after a limit number of steps. This is because there is a natural variance in the types of mathematical entities that get constructed, and calls to the Z3 theorem prover can often take many seconds. When resolving conjectures, we set the timeout to Z3 to be $2.0$ seconds, and to $0.5$ seconds when using Z3 to add instances to entities without any.

\subsection{REPL}

\fermat also comes equipped with a read-eval-print-loop (REPL) for manual interaction with the environment. The REPL allows the user to use an interactive shell to define, inspect, and evaluate mathematical entities. 

\paragraph{Available commands.}

\begin{center}
    \begin{tabular}{@{}ll@{}}
        \toprule
        \textbf{Command} & \textbf{Description}\\
        \midrule
        \texttt{help}      & Get help on commands or list available rules\\
        \texttt{list}      & List available concepts, rules, or conjectures\\
        \texttt{apply}     & Apply a production rule to create new concepts/conjectures\\
        \texttt{inspect}   & Show detailed information about an entity\\
        \texttt{compute}   & Test computational implementation with arguments\\
        \texttt{rename}    & Rename an entity\\
        \texttt{remove}    & Remove an entity\\
        \texttt{visualize} & Create a visualization of the current knowledge graph\\
        \texttt{clear}     & Clear the screen\\
        \texttt{save}      & Save knowledge graph to file \\
        \texttt{exit}      & Exit the REPL\\
        \bottomrule
    \end{tabular}
\end{center}

\paragraph{Example usage.}

\begin{center}
\begin{tabular}{@{}ll@{}}
\toprule
\textbf{REPL Command} & \textbf{Concept Produced}\\
\midrule
\texttt{apply iter successor} &
\emph{add}\\
\texttt{apply iter add zero} &
\emph{multiply}\\
\texttt{apply match add indices\_to\_match=[0,1]} &
\emph{double}\\
\texttt{apply match multiply indices\_to\_match=[0,1]} &
\emph{square}\\
\texttt{apply specialize successor zero index\_to\_specialize=0} &
\emph{one}\\
\bottomrule
\end{tabular}  
\end{center}

\begin{figure}
\begin{mdframed}[roundcorner=10pt]
\begin{minted}[
    breaklines,
    breaksymbolleft={},
    % breaksymbolright={},
    % breaksymbol={},
    fontsize=\small
]{md}
Write a Python function called `calculate_interestingness` that takes an entity_id and a knowledge graph, and returns a float between 0 and 1 representing how interesting the entity is. Use the primitive functions to extract relevant features and combine them in a meaningful way.

  Your function should aim to identify entities that are:
  1. Not too simple and not overtly complicated and uninteresting.
  2. Likely to be fruitful for further exploration
  3. Have characteristics that mathematicians would find interesting
  4. Possess a good balance of generality and specificity

  You can use mathematical operations (arithmetic, min/max, etc.) and common Python libraries like math and numpy to combine the primitives in a way that captures interestingness. CONSTRAINTS:
  - You **MUST** respond with **only** the complete, syntactically correct Python code for the new function (`calculate_interestingness_vN`).
  - Include the `def calculate_interestingness_vN(...):` signature line and the function body. Add a concise docstring.
  - **DO NOT** include any introductory text, explanations, comments outside the function body, or usage examples in your response.
  - If you use any of the primitives or abstractions, make sure you use them correctly. Provide the right inputs as described in the documentation given about the primitives!
  - The descriptions of the primitives and abstractions indicate what arguments they take. Follow proper Python syntax. Watch out for potential division by zero errors.

You have access to the following primitive functions that can be used in your interestingness function. Each primitive provides some information about a mathematical entity in the knowledge graph: {}

  Please implement the interestingness measure in the following format:

      def calculate_interestingness(entity_id: str, graph: KnowledgeGraph) -> float:
        """
        Calculate the interestingness of a mathematical entity.
        
        Args:
            entity_id: The ID of the entity in the knowledge graph
            graph: The knowledge graph containing all mathematical entities
            
        Returns:
            A float between 0 and 1 representing how interesting the entity is,
            where 0 is least interesting and 1 is most interesting
        """
        # Implement your interestingness scoring here
        # ...
        
        # Return a value between 0 and 1
        return score

Make sure your function handles potential errors gracefully, for example by catching exceptions when calling primitives. The function should always return a valid float between 0 and 1, even if there are unexpected inputs or errors.
\end{minted}
\end{mdframed}
\caption{The one-shot prompt for our GPT-$4$o baseline. We do not insert the primitives here for brevity, these can be found in Figure \ref{fig:primitives}.}
\label{fig:one-shot-prompt}
\end{figure}

\begin{figure}
\begin{mdframed}[roundcorner=10pt]
\begin{minted}[
    breaklines,
    breaksymbolleft={},
    % breaksymbolright={},
    % breaksymbol={},
    fontsize=\small
]{md}
- `get_ancestors(entity_id, graph)`: Returns list of ancestor node IDs.
- `get_descendants(entity_id, graph)`: Returns list of descendant node IDs.
- `get_construction_depth(entity_id, graph)`: Returns the longest path from a root node.
- `get_in_degree(entity_id, graph)`: Returns the number of direct parent nodes.
- `get_out_degree(entity_id, graph)`: Returns the number of direct child nodes.
- `get_construction_history_rule_names(entity_id, graph)`: Returns list of rule names used in construction.
- `get_entity_step_age(entity_id, graph)`: Returns the entity's age in construction steps.
- `get_num_concepts(graph)`: Returns the total number of concepts.
- `get_num_conjectures(graph)`: Returns the total number of conjectures.
- `get_entity_node_type(entity_id, graph)`: Returns 'Concept', 'Conjecture', or 'Theorem' depending on the type of the entity.
- `get_concept_category(entity_id, graph)`: Returns 'Predicate', 'Function', or 'Constant' depending on the type of the entity.
- `get_input_arity(entity_id, graph)`: Returns input arity of the entity..
- `get_num_component_types(entity_id, graph)`: Returns number of component types in examples.
- `get_examples(entity_id, graph)`: Returns list of positive examples, each example is a tuple of ints.
- `get_nonexamples(entity_id, graph)`: Returns list of negative examples, each example is a tuple of ints.
- `get_num_construction_inputs(entity_id, graph)`: Returns number of direct construction inputs.
- `is_proven(entity_id, graph)`: Returns 1.0 if proven theorem, 0.0 otherwise.
- `create_weighted_interestingness_function(functions: List[Callable], weights: List[float])`: Creates a weighted interestingness function from a list of interestingness functions and a list of weights.
\end{minted}
\end{mdframed}
\caption{A list of primitive methods available to the interestingness measure synthesizers. Each method returns a simple property or information about the knowledge graph and/or the input entity.}
\label{fig:primitives}
\end{figure}

\begin{figure}
\begin{mdframed}[roundcorner=10pt]
\begin{minted}[
    breaklines,
    breaksymbolleft={},
    % breaksymbolright={},
    % breaksymbol={},
    fontsize=\small
]{md}
You are an expert Python programming assistant specializing in evolving code based on performance feedback.

You are participating in an evolutionary function discovery process (FunSearch) to find a high-performing Python function called `calculate_interestingness`. This function evaluates the 'interestingness' of mathematical entities within a knowledge graph.

The user prompt contains several **example implementations** of this function (named `calculate_interestingness_v0`, `calculate_interestingness_v1`, etc.), showcasing different approaches that have shown some success. The prompt ends with the header for the **new function** you need to generate: `def calculate_interestingness_vN(entity_id: str, graph: KnowledgeGraph) -> float:`.

Your specific task is to **generate a new, potentially improved version** of the `calculate_interestingness` function, named `calculate_interestingness_vN`. You should **analyze all the example functions** provided in the user prompt (`_v0` to `_v(N-1)`) to understand different successful strategies and potentially combine or adapt their ideas.

The function you write will receive `entity_id` (string) and `graph` (a `KnowledgeGraph` object) as input. You can use the following methods on the `graph` object to get information about the entity (`entity_id`) or the graph itself, the description explains what arguments it takes: {}

Optionally, you can also use the following abstractions: {}

You also have access to standard Python libraries like `math`. Do not use notation like `graph.METHOD_NAME(args)`, only `METHOD_NAME(args)` will work.

The goal is to create a function that receives a higher score when evaluated, indicating it better captures mathematical interestingness.

**Output Constraints:**
- You **MUST** respond with **only** the complete, syntactically correct Python code for the new function (`calculate_interestingness_vN`).
- Include the `def calculate_interestingness_vN(...):` signature line and the function body. Add a concise docstring.
- **DO NOT** include any introductory text, explanations, comments outside the function body, or usage examples in your response.
- Enclose the entire function definition within a single markdown code block like this:
- If you use any of the primitives or abstractions, make sure you use them correctly by supplying the proper arguments. 
- Try not to rely on the abstractions alone - use them in a compositional way, where you also implement some of the logic yourself (passing interesting arguments to the abstractions counts).
- Try not to copy the examples exactly, but rather use them as inspiration to create a new, better, function that *can* be similar.
- You do not have to use all primitives, and you do not have to make extremely complex functions if you don't think it necessary.
- Watch out for potential division by zero errors.

```python
def calculate_interestingness_vN(entity_id: str, graph: KnowledgeGraph) -> float:
    """A new function version inspired by provided examples."""
    # ... implementation ...
    return score
```
\end{minted}
\end{mdframed}
\caption{The prompt supplied to the evolution sampler $\mathcal{L}_{var}$, indicating the evolution task that needs to be applied. We have removed the description of the DSL primitives which appears in \ref{fig:primitives}.}
\label{fig:evolution_prompt}
\end{figure}

\begin{figure}
\begin{mdframed}[roundcorner=10pt]
\begin{minted}[
    breaklines,
    breaksymbolleft={},
    % breaksymbolright={},
    % breaksymbol={},
    fontsize=\small
]{md}
  You are an expert programmer specializing in code refactoring and identifying useful, general-purpose abstractions within existing code.
  You will be given a set of Python functions, each with a performance score, and a list of already-identified abstractions. Your task is to analyze the functions and extract new, reusable subroutines.

  An "abstraction" is a self-contained function that performs a useful calculation. It should be general enough to be used in various contexts. For example, an abstraction can generalize a pattern by turning constants into parameters.

  You must only return the Python code blocks for the new abstractions you create. Do not include any other text, explanation, or conversation.

  Your goal is to identify and implement useful, reusable subroutines (abstractions) from the provided program examples.

  ## 1. Existing Abstractions
  Review the following abstractions that have already been created. **Avoid creating new abstractions that are functionally identical to these.**
  {current_abstractions}

  ## 2. Program Examples to Analyze
  Here are the programs to analyze, along with their performance scores. You should **prioritize creating abstractions from programs with higher scores**, as they are more likely to contain useful logic.
  {program_examples}

  ## 3. Your Task & Guiding Principles
  Carefully analyze the program examples and identify common or useful patterns that can be generalized into new abstractions.
  - An abstraction should be a **small, reusable function** that captures a specific calculation or logical step.
  - Good abstractions are **general**. Instead of hard-coding values, define them as function arguments. For example, if you see `(x - y) * 0.5` in a program, a good abstraction would be `def scaled_difference(a, b, factor): return (a - b) * factor`, not `def specific_difference(a, b): return (a - b) * 0.5`.
  - You can create **improved or generalized versions** of existing abstractions. If you do, append `_v2`, `_v3`, etc., to the original name to ensure it is unique.
  - You may also **compose existing abstractions** to create a new, more powerful one.

  ## 4. Required Output Format
  Provide your response as a list of Python functions. Each function must have a concise docstring explaining its purpose. Use descriptive argument names.

  ```python
  def new_abstraction_name(arg1, arg2: float) -> any:
      """
      A concise description of what this abstraction calculates.
      """
      # ... implementation ...
      return result
\end{minted}
\end{mdframed}
\caption{The prompt supplied to the abstraction sampler $\mathcal{L}_{abs}$, indicating the abstraction task that needs to be carried out.}
\label{fig:abstraction_prompt}
\end{figure}

\begin{figure}
\begin{mdframed}[roundcorner=10pt]
\begin{minted}[
    breaklines,
    breaksymbolleft={},
    % breaksymbolright={},
    % breaksymbol={},
    fontsize=\small
]{md}
def calculate_interestingness(entity_id: str, graph) -> float:
    """
    Calculate the interestingness of a mathematical entity.
    
    Args:
        entity_id: The ID of the entity in the knowledge graph
        graph: The knowledge graph containing all mathematical entities
        
    Returns:
        A float between 0 and 1 representing how interesting the entity is,
        where 0 is least interesting and 1 is most interesting
    """
    import numpy as np

    try:
        # Get various properties of the entity
        construction_depth = get_construction_depth(entity_id, graph)
        in_degree = get_in_degree(entity_id, graph)
        out_degree = get_out_degree(entity_id, graph)
        num_construction_inputs = get_num_construction_inputs(entity_id, graph)
        node_type = get_entity_node_type(entity_id, graph)
        input_arity = get_input_arity(entity_id, graph)
        num_component_types = get_num_component_types(entity_id, graph)
        is_proven_theorem = is_proven(entity_id, graph)
        
        # Calculate base metrics with small adjustments to prevent division by zero
        complexity = np.log(1 + construction_depth) / (1 + input_arity)
        influence = (in_degree * out_degree) / (1 + num_construction_inputs)
        specificity = 1 / (1 + num_component_types)
        proof_bonus = 0.1 if is_proven_theorem else 0
        
        # Combine metrics with weights; weights can be adjusted as needed
        score = (0.4 * complexity + 0.3 * influence + 0.2 * specificity + 0.1 * proof_bonus)
        
        # Ensure score is between 0 and 1
        return min(max(score, 0.0), 1.0)

    except Exception as e:
        # Handle unexpected errors by returning a neutral score
        return 0.5
\end{minted}
\end{mdframed}
\caption{An interestingness measure generated by GPT-4o. It begins by extracting relevant features of the state using the primitives. The measure itself is not very performant as it overly rewards complexity and node connectivity in the graph, which only increase in new entities. As ground-truth entities are not developed immediately using this measure, the episodes proceed by producing increasingly convoluted and uninteresting objects.}
\label{fig:gpt-4o-measure}
\end{figure}

\begin{figure}
\begin{mdframed}[roundcorner=10pt]
\begin{minted}[
    breaklines,
    breaksymbolleft={},
    % breaksymbolright={},
    % breaksymbol={},
    fontsize=\tiny
]{md}
def calculate_interestingness(entity_id: str, graph: KnowledgeGraph) -> float:
  """Calculate the interestingness score for a given entity.
    
    Args:
        entity_id: The ID of the entity to score.
        graph: The knowledge graph containing the entity.
        
    Returns:
        A float value representing the interestingness score (higher is more interesting).
    """
  try:
      # Retrieve entity metrics
      metrics = retrieve_entity_metrics(entity_id, graph)
      (node_type, depth, in_degree, out_degree, step_age, num_concepts, 
       num_conjectures, proven_status, arity, num_components, 
       num_construction_inputs, ancestors_count, descendants_count, 
       examples, nonexamples, rules) = metrics

      # Base score calculations
      (depth_score, connectivity_score, age_score, arity_score) = calculate_base_scores(
          depth, in_degree, out_degree, step_age, arity, num_components, num_concepts, num_conjectures)

      # Calculate additional influence scores
      influence_score = calculate_influence_score(ancestors_count, descendants_count, num_concepts, num_conjectures)

      # Calculate example-based scores
      example_balance = calculate_example_balance(examples, nonexamples)
      uniqueness_score = calculate_uniqueness_score_v2(examples, nonexamples)

      # Rule diversity score
      rule_diversity_score = calculate_rule_diversity_score(rules)

      # Complexity score
      complexity_score = calculate_complexity_score(num_construction_inputs, depth)

      # Weights and score calculation
      weights = {
          'depth': 0.15,
          'connectivity': 0.15,
          'age': 0.1,
          'arity': 0.1,
          'proven_status': 0.1,
          'influence': 0.2,
          'example_balance': 0.05,
          'uniqueness': 0.1,
          'rule_diversity': 0.05,
          'complexity': 0.1
      }

      # Calculate overall score
      score = calculate_combined_score(
          depth_score, connectivity_score, age_score, arity_score, 
          proven_status, influence_score, example_balance, uniqueness_score,
          rule_diversity_score, complexity_score, category_bonus=0.0, weights=weights)

      # Adjust score by node type with refined multipliers
      score = adjust_score_by_node_type(score, node_type, concept_multiplier=1.3, conjecture_multiplier=1.2)

      return score

  except Exception:
      return 0.0
    
\end{minted}
\end{mdframed}
\caption{The best program found by EvoAbstract in our main run on the starting knowledge graph \texttt{succ\_zero\_eq}. We include the abstractions identified which are used in this program in Figure \ref{fig:evoabstract-succ-abstractions}.}
\label{fig:evoabstract-best-succ}
\end{figure}

\begin{figure}
\begin{mdframed}[roundcorner=10pt]
\begin{minted}[
    breaklines,
    breaksymbolleft={},
    % breaksymbolright={},
    % breaksymbol={},
    fontsize=\tiny
]{md}
def retrieve_entity_metrics(entity_id: str, graph: KnowledgeGraph) -> tuple:
    node_type = get_entity_node_type(entity_id, graph)
    depth = get_construction_depth(entity_id, graph)
    in_degree = get_in_degree(entity_id, graph)
    out_degree = get_out_degree(entity_id, graph)
    step_age = get_entity_step_age(entity_id, graph)
    num_concepts = get_num_concepts(graph)
    num_conjectures = get_num_conjectures(graph)
    proven_status = is_proven(entity_id, graph)
    arity = get_input_arity(entity_id, graph)
    num_components = get_num_component_types(entity_id, graph)
    num_construction_inputs = get_num_construction_inputs(entity_id, graph)
    ancestors_count = len(get_ancestors(entity_id, graph))
    descendants_count = len(get_descendants(entity_id, graph))
    examples = get_examples(entity_id, graph)
    nonexamples = get_nonexamples(entity_id, graph)
    rules = get_construction_history_rule_names(entity_id, graph)
    
    return (node_type, depth, in_degree, out_degree, step_age, num_concepts, num_conjectures, proven_status, arity, num_components, num_construction_inputs, ancestors_count, descendants_count, examples, nonexamples, rules)

def calculate_base_scores(depth, in_degree, out_degree, step_age, arity, num_components, num_concepts, num_conjectures) -> tuple:

    depth_score = depth / (1 + num_concepts)
    connectivity_score = (in_degree + out_degree) / (1 + num_concepts + num_conjectures)
    age_score = step_age / (1 + depth)
    arity_score = arity / (1 + num_components)
    return depth_score, connectivity_score, age_score, arity_score

def calculate_influence_score(ancestors_count, descendants_count, num_concepts, num_conjectures) -> float:
    """
    Calculate influence score considering ancestors and descendants.

    Returns the influence score based on ancestor and descendant counts.
    """
    return (ancestors_count + descendants_count) / (1 + num_concepts + num_conjectures)

def calculate_example_balance(examples, nonexamples) -> float:
    """
    Calculate the example balance score.

    This score reflects the proportion of examples compared to nonexamples,
    with an adjustment to prevent division by zero.
    """
    return len(examples) / (1 + len(nonexamples))

def calculate_uniqueness_score_v2(examples, nonexamples) -> float:
    """
    Calculate uniqueness score based on the difference between example and non-example sets.
    This abstraction is an improvement on calculate_uniqueness_score, allowing flexibility in weighting the size of examples.
    """
    unique_examples_count = len(set(examples).difference(set(nonexamples)))
    return unique_examples_count / (1 + len(examples))

def calculate_rule_diversity_score(rules) -> float:
    """
    Calculate rule diversity score for a set of rules.
    
    Returns the diversity score based on the uniqueness of construction rules.
    """
    return len(set(rules)) / (1 + len(rules))

def calculate_complexity_score(construction_inputs, depth) -> float:
    """
    Calculate the complexity score based on construction inputs and depth.
    """
    return construction_inputs / (1 + depth)

def calculate_combined_score(depth_score, connectivity_score, age_score, arity_score, proven_status, influence_score, example_balance, uniqueness_score, rule_diversity_score=None, input_diversity_score=None, category_bonus=0.0, weights=None) -> float:
    """
    Combine various component scores into a single score, accounting for weights and optional inputs. Computes a weighted sum.
    """
    {implementation removed for brevity}
    return score

def adjust_score_by_node_type(score, node_type, concept_multiplier=1.2, conjecture_multiplier=1.1) -> float:
    if node_type == 'Concept':
        return score * concept_multiplier
    elif node_type == 'Conjecture':
        return score * conjecture_multiplier
    return score
\end{minted}
\end{mdframed}
\caption{The abstractions used in the best program found by EvoAbstract during the run on \texttt{succ\_zero\_eq}.}
\label{fig:evoabstract-succ-abstractions}
\end{figure}

\begin{figure}
\begin{mdframed}[roundcorner=10pt]
\begin{minted}[
    breaklines,
    breaksymbolleft={},
    % breaksymbolright={},
    % breaksymbol={},
    fontsize=\tiny
]{md}
def calculate_interestingness_v2(entity_id: str, graph: KnowledgeGraph) -> float:
    """Improved version of `calculate_interestingness_v1` with enhanced score integration."""
    # Retrieve entity characteristics
    node_type, concept_category, in_degree, out_degree, construction_depth, step_age, num_construction_inputs, proven_status = retrieve_entity_characteristics(entity_id, graph)

    # Base score calculation
    base_score = compute_base_score(node_type, concept_category)

    # Diversity score using construction rule names
    rule_names = get_construction_history_rule_names(entity_id, graph)
    diversity_score = calculate_diversity_score_v2(rule_names, factor=0.18)  # Balanced diversity factor

    # Connectivity score with adjusted factor
    connectivity_score = calculate_comprehensive_connectivity_score(in_degree, out_degree, construction_depth, factor=1.1)

    # Construction score with emphasis on age
    construction_score = calculate_emphasized_construction_score(step_age, num_construction_inputs, age_factor=1.2, input_factor=0.4)

    # Create weights dictionary
    weights = {
        'base': 1.2,             # Increased emphasis on base
        'connectivity': 0.85,    # Higher connectivity weight
        'construction': 0.95,    # Slightly reduced construction weight
        'diversity': 0.25,       # Consistent diversity emphasis
        'boost_amount': 0.9      # Slightly reduced boost for proven status
    }

    # Compute the final score
    score = compute_final_score(base_score, connectivity_score, construction_score, diversity_score, proven_status == 1.0, weights)

    return score

\end{minted}
\end{mdframed}
\caption{A performant program found by EvoAbstract in our main run on the starting knowledge graph \texttt{ff\_27}. We include the abstractions identified which are used in this program in Figure \ref{fig:evoabstract-ff-abstractions}.}
\label{fig:evoabstract-best-ff}
\end{figure}

\begin{figure}
\captionsetup{aboveskip=0pt,belowskip=0.3pt}
\caption{The abstractions used in the best program found by EvoAbstract during the run on \texttt{ff\_27}. Note that often newer abstractions use previous ones, sometimes trivially.}
\begin{mdframed}[roundcorner=10pt]
\begin{minted}[
    breaklines,
    breaksymbolleft={},
    % breaksymbolright={},
    % breaksymbol={},
    fontsize=\tiny,
]{md}

def calculate_comprehensive_connectivity_score(in_degree: int, out_degree: int, construction_depth: int, factor: float) -> float:
    """
    Calculate the comprehensive connectivity score of an entity.

    Args:
        in_degree: The in-degree of the entity.
        out_degree: The out-degree of the entity.
        construction_depth: The construction depth of the entity.
        factor: The scaling factor for the connectivity score.

    Returns:
        A float value representing the comprehensive connectivity score.
    """
    return calculate_connectivity_score_with_factors(in_degree, out_degree, construction_depth, factor)


def calculate_emphasized_construction_score(step_age: int, num_construction_inputs: int, age_factor: float, input_factor: float) -> float:
    """
    Calculate an emphasized construction score based on step age and number of construction inputs,
    adjusted by specified age and input factors.

    Args:
        step_age: The age of the construction step.
        num_construction_inputs: The number of inputs in the construction.
        age_factor: The factor to adjust the emphasis on the age of the step.
        input_factor: The factor to adjust the emphasis on the number of inputs.

    Returns:
        A float value representing the emphasized construction score.
    """
    return (step_age * age_factor) / (num_construction_inputs * input_factor + 1)


def calculate_base_score(node_type: str, concept_category: str) -> float:
    """
    Calculate the base score for an entity based on its node type and concept category.
    """
    base_score = 0.0
    if node_type == 'Theorem':
        base_score += 1.0
    elif node_type == 'Conjecture':
        base_score += 0.8
    else:
        base_score += 0.5

    if concept_category == 'Function':
        base_score += 0.3
    elif concept_category == 'Predicate':
        base_score += 0.2
    
    return base_score

def calculate_connectivity_score_with_factors(in_degree: int, out_degree: int, construction_depth: int, factor: float = 1.0) -> float:
    return ((in_degree + out_degree) / (construction_depth + 1)) * factor


def calculate_diversity_score_v2(rule_names: list, factor: float) -> float:
    """
    Calculate the diversity score based on the unique rule names and a factor.

    Args:
        rule_names: A list of rule names associated with construction history.
        factor: A float value representing the factor to scale the diversity.

    Returns:
        A float value representing the diversity score.
    """
    return len(set(rule_names)) * factor

def compute_final_score(base_score: float, connectivity_score: float, construction_score: float, diversity_score: float, proven_status: bool, weights: dict) -> float:
    """
    Compute the final interestingness score after combining scores and potentially boosting for proven status.

    Args:
        base_score: The base component of the score.
        connectivity_score: The connectivity component of the score.
        construction_score: The construction component of the score.
        diversity_score: The diversity component of the score.
        proven_status: Whether the entity is proven or not.
        weights: A dictionary containing weights and a boost amount for proven status.

    Returns:
        A float representing the final computed score.
    """
    score = calculate_scores_v2(base_score, connectivity_score, construction_score, diversity_score, weights)
    return adjust_score_for_status_v2(score, proven_status, weights['boost_amount'])
\end{minted}
\end{mdframed}
\label{fig:evoabstract-ff-abstractions}
\end{figure}

\begin{figure}
\begin{mdframed}[roundcorner=10pt]
\begin{minted}[
    breaklines,
    breaksymbolleft={},
    % breaksymbolright={},
    % breaksymbol={},
    fontsize=\tiny
]{md}
def calculate_interestingness(entity_id: str, graph: KnowledgeGraph) -> float:
  """Calculate the interestingness score for a given entity.
    
    Args:
        entity_id: The ID of the entity to score.
        graph: The knowledge graph containing the entity.
        
    Returns:
        A float value representing the interestingness score (higher is more interesting).
    """
  try:
      # Retrieve properties of the entity
      in_degree = get_in_degree(entity_id, graph)
      out_degree = get_out_degree(entity_id, graph)
      construction_depth = get_construction_depth(entity_id, graph)
      entity_step_age = get_entity_step_age(entity_id, graph)
      node_type = get_entity_node_type(entity_id, graph)
      num_examples = len(get_examples(entity_id, graph))
      num_nonexamples = len(get_nonexamples(entity_id, graph))
      num_concepts = get_num_concepts(graph)
      num_conjectures = get_num_conjectures(graph)
      input_arity = get_input_arity(entity_id, graph)
      num_component_types = get_num_component_types(entity_id, graph)
      num_construction_inputs = get_num_construction_inputs(entity_id, graph)

      # Calculate structural score with construction depth normalization
      structural_score = (in_degree + out_degree + math.log1p(num_examples + num_nonexamples)) / (1 + math.sqrt(construction_depth) + num_construction_inputs)

      # Enhance novelty score using exponential decay
      novelty_score = math.exp(-0.05 * entity_step_age)

      # Calculate density factor taking root of concept and conjecture counts
      density_factor = (math.sqrt(num_concepts) + math.sqrt(num_conjectures)) / (1 + construction_depth) if (num_concepts + num_conjectures) > 0 else 1.0

      # Calculate type score with increased weight for proven theorems
      type_score = 0.0
      if node_type == 'Concept':
          type_score = 1.2 * density_factor
      elif node_type == 'Conjecture':
          type_score = 0.7 * density_factor
      elif node_type == 'Theorem':
          type_score = is_proven(entity_id, graph) * 1.8 * density_factor

      # Arity complexity with stronger penalty for multiple types
      arity_complexity_factor = 1.0 / (1 + 2 * input_arity + num_component_types)

      # Calculate balance in examples to non-examples with log adjustment
      example_balance_score = (2 * math.log1p(num_examples) + math.log1p(num_nonexamples)) / math.log1p(num_examples + num_nonexamples + 1)

      # Example ratio to emphasize more positive examples
      example_ratio = (num_examples - num_nonexamples) / (1 + num_examples + num_nonexamples)

      # Aggregate score with adjusted weight distribution
      score = (
          0.3 * structural_score +
          0.2 * novelty_score * arity_complexity_factor +
          0.2 * type_score * arity_complexity_factor +
          0.15 * example_balance_score +
          0.1 * (1.0 / (1 + num_component_types)) +
          0.05 * example_ratio
      )
      return score

  except Exception:
      return 0.0

\end{minted}
\end{mdframed}
\caption{The best program found by FunSearch during the run on \texttt{succ\_zero\_eq}.}
\label{fig:funsearch-best-succ}
\end{figure}

\begin{figure}
    \centering
    \caption{Sample sections of elementary number theory discovered by EvoAbstract during runs on \texttt{succ\_zero\_eq} and \texttt{arith\_base}.}
    \includegraphics[width=1.07\linewidth]{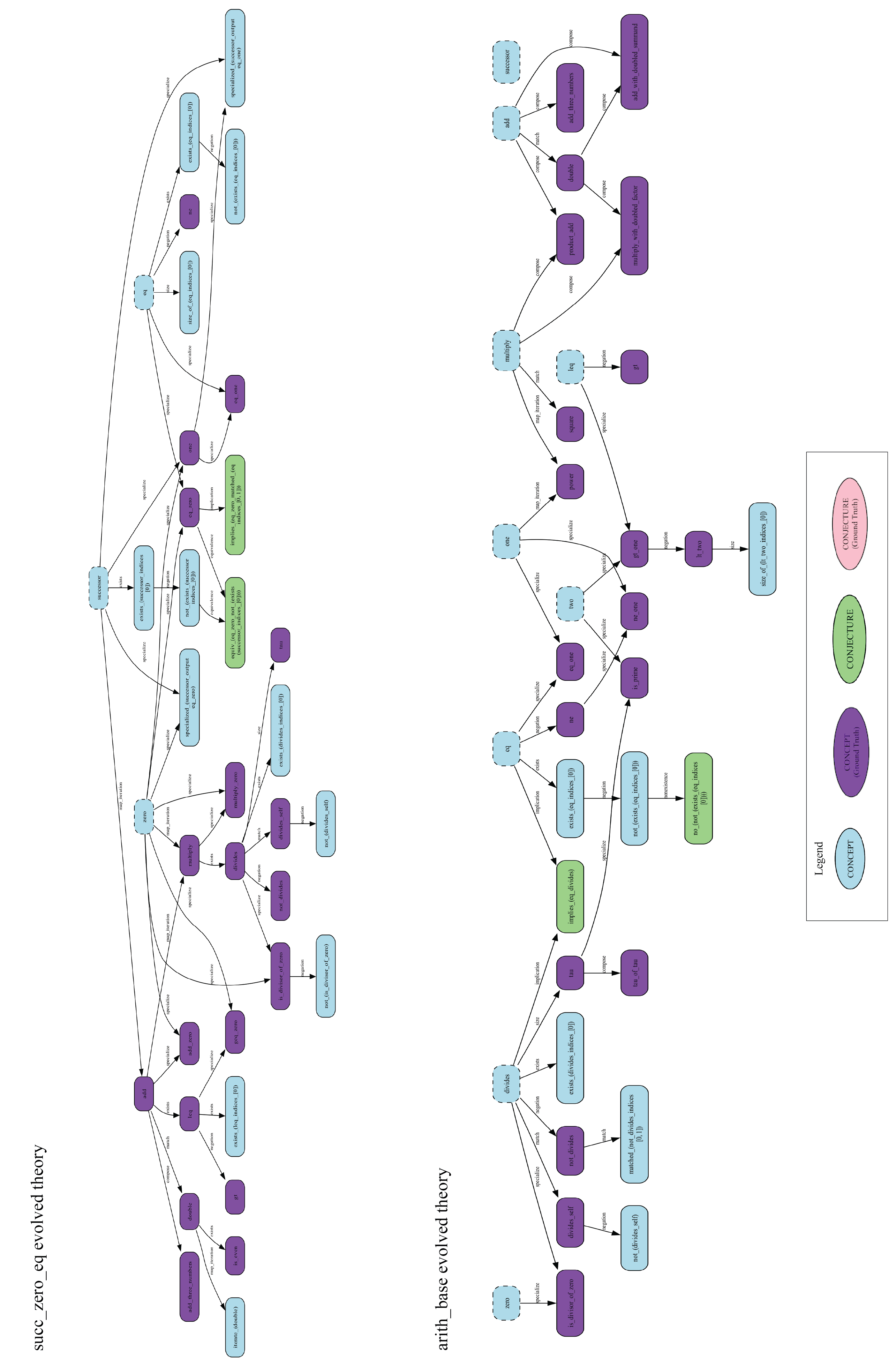}

\label{fig:learnedgraphs}
\end{figure}

% ----------------------------------------------------------------

%%%%%%%%%%%%%%%%%%%%%%%%%%%%%%%%%%%%%%%%%%%%%%%%%%%%%%%%%%%%

\clearpage
\section*{NeurIPS Paper Checklist}

\begin{enumerate}

\item {\bf Claims}
    \item[] Question: Do the main claims made in the abstract and introduction accurately reflect the paper's contributions and scope?
    \item[] Answer: \answerYes{} % Replace by \answerYes{}, \answerNo{}, or \answerNA{}.
    \item[] Justification: Yes, the abstract and introductory aptly describe our contributions in \textsc{Fermat}, investigations in interestingness learning, and EvoAbstract.

\item {\bf Limitations}
    \item[] Question: Does the paper discuss the limitations of the work performed by the authors?
    \item[] Answer: \answerYes{} % Replace by \answerYes{}, \answerNo{}, or \answerNA{}.
    \item[] Justification: Yes, our paper discusses limitations of our investigation and framework in the discussion portion of the experiments section (Section 5).

\item {\bf Theory assumptions and proofs}
    \item[] Question: For each theoretical result, does the paper provide the full set of assumptions and a complete (and correct) proof?
    \item[] Answer: \answerNA{} % Replace by \answerYes{}, \answerNo{}, or \answerNA{}.
    \item[] Justification: Our paper does not include theoretical results.

    \item {\bf Experimental result reproducibility}
    \item[] Question: Does the paper fully disclose all the information needed to reproduce the main experimental results of the paper to the extent that it affects the main claims and/or conclusions of the paper (regardless of whether the code and data are provided or not)?
    \item[] Answer: \answerYes{} % Replace by \answerYes{}, \answerNo{}, or \answerNA{}.
    \item[] Justification: Yes, the paper discloses all information required to reproduce the main results.

\item {\bf Open access to data and code}
    \item[] Question: Does the paper provide open access to the data and code, with sufficient instructions to faithfully reproduce the main experimental results, as described in supplemental material?
    \item[] Answer: \answerYes{} % Replace by \answerYes{}, \answerNo{}, or \answerNA{}.
    \item[] Justification: Yes, we include our code and commands for running experiments in the supplementary material.

\item {\bf Experimental setting/details}
    \item[] Question: Does the paper specify all the training and test details (e.g., data splits, hyperparameters, how they were chosen, type of optimizer, etc.) necessary to understand the results?
    \item[] Answer: \answerYes{} % Replace by \answerYes{}, \answerNo{}, or \answerNA{}.
    \item[] Justification: Yes, we share all details for hyperparameters in our experiments.

\item {\bf Experiment statistical significance}
    \item[] Question: Does the paper report error bars suitably and correctly defined or other appropriate information about the statistical significance of the experiments?
    \item[] Answer: \answerYes{} % Replace by \answerYes{}, \answerNo{}, or \answerNA{}.
    \item[] Justification: Yes, we report standard deviations of rewards obtained during experiments with different measures, and run our method averaged over 4 runs for a comparison with FunSearch.

\item {\bf Experiments compute resources}
    \item[] Question: For each experiment, does the paper provide sufficient information on the computer resources (type of compute workers, memory, time of execution) needed to reproduce the experiments?
    \item[] Answer: \answerYes{} % Replace by \answerYes{}, \answerNo{}, or \answerNA{}.
    \item[] Justification: We list the computational resources we used in the Appendix.
    
\item {\bf Code of ethics}
    \item[] Question: Does the research conducted in the paper conform, in every respect, with the NeurIPS Code of Ethics \url{https://neurips.cc/public/EthicsGuidelines}?
    \item[] Answer: \answerYes{} % Replace by \answerYes{}, \answerNo{}, or \answerNA{}.
    \item[] Justification: We have verified that our research conforms with the NeurIPS code of ethics.

\item {\bf Broader impacts}
    \item[] Question: Does the paper discuss both potential positive societal impacts and negative societal impacts of the work performed?
    \item[] Answer: \answerNA{} % Replace by \answerYes{}, \answerNo{}, or \answerNA{}.
    \item[] Justification: There is no immediate societal impact of the work performed.
    
\item {\bf Safeguards}
    \item[] Question: Does the paper describe safeguards that have been put in place for responsible release of data or models that have a high risk for misuse (e.g., pretrained language models, image generators, or scraped datasets)?
    \item[] Answer: \answerNA % Replace by \answerYes{}, \answerNo{}, or \answerNA{}.
    \item[] Justification: Our paper poses no such risks.

\item {\bf Licenses for existing assets}
    \item[] Question: Are the creators or original owners of assets (e.g., code, data, models), used in the paper, properly credited and are the license and terms of use explicitly mentioned and properly respected?
    \item[] Answer: \answerYes{} % Replace by \answerYes{}, \answerNo{}, or \answerNA{}.
    \item[] Justification: All assets we do not exclusively develop are properly credited and respected.

\item {\bf New assets}
    \item[] Question: Are new assets introduced in the paper well documented and is the documentation provided alongside the assets?
    \item[] Answer: \answerYes{} % Replace by \answerYes{}, \answerNo{}, or \answerNA{}.
    \item[] Justification: Yes, our new framework and codebase comes with documentation.

\item {\bf Crowdsourcing and research with human subjects}
    \item[] Question: For crowdsourcing experiments and research with human subjects, does the paper include the full text of instructions given to participants and screenshots, if applicable, as well as details about compensation (if any)? 
    \item[] Answer: \answerNA{} % Replace by \answerYes{}, \answerNo{}, or \answerNA{}.
    \item[] Justification: Our paper does not involve crowdsourcing nor research with human subjects.

\item {\bf Institutional review board (IRB) approvals or equivalent for research with human subjects}
    \item[] Question: Does the paper describe potential risks incurred by study participants, whether such risks were disclosed to the subjects, and whether Institutional Review Board (IRB) approvals (or an equivalent approval/review based on the requirements of your country or institution) were obtained?
    \item[] Answer: \answerNA{} % Replace by \answerYes{}, \answerNo{}, or \answerNA{}.
    \item[] Justification: Our paper does not involve crowdsourcing nor research with human subjects.

\item {\bf Declaration of LLM usage}
    \item[] Question: Does the paper describe the usage of LLMs if it is an important, original, or non-standard component of the core methods in this research? Note that if the LLM is used only for writing, editing, or formatting purposes and does not impact the core methodology, scientific rigorousness, or originality of the research, declaration is not required.
    %this research? 
    \item[] Answer: \answerNA{} % Replace by \answerYes{}, \answerNo{}, or \answerNA{}.
    \item[] Justification: We did not use LLMs to conduct any core method development in this paper.

\end{enumerate}

\end{document}